\documentclass[10pt,twocolumn,letterpaper]{article}

\usepackage{cvpr}              
\definecolor{cvprblue}{rgb}{0.21,0.49,0.74}
\usepackage[pagebackref,breaklinks,colorlinks,allcolors=cvprblue]{hyperref}

\usepackage{color}
\usepackage{xcolor}
\usepackage{colortbl}
\definecolor{Gray}{gray}{0.9}
\usepackage{threeparttable}
\usepackage{multirow}
\usepackage{array}
\usepackage{makecell}
\newcolumntype{C}[1]{>{\centering}p{#1}}
\usepackage{epsfig}
\usepackage{graphicx}
\usepackage{float}
\definecolor{ours}{RGB}{244,237,252}

\def\paperID{xxxx} 
\def\confName{CVPR}
\def\confYear{2026}

\title{WeDetect: Fast Open-Vocabulary Object Detection as Retrieval}

\author{
Shenghao Fu$^{1\dag}$, Yukun Su$^{1\dag}$, Fengyun Rao$^{1*}$, Jing LYU$^{1}$, 
{Xiaohua Xie$^{2}$\thanks{: Corresponding authors are Xiaohua Xie and Fengyun Rao.\\\indent $^{\dag}$: Equal Contribution.\\\indent Work was done when Shenghao Fu was an intern at Tencent. }, Wei-Shi Zheng$^{2,3}$} \\
{\small $^1$WeChat Vision, Tencent Inc. \quad $^2$Pazhou Laboratory (Huangpu), China \quad $^3$Shenzhen Loop Area Institute} \\
{\small 1185514120@qq.com, \{yukunsu,fengyunrao\}@tencent.com, wszheng@ieee.org}
}

\begin{document}
\maketitle

\begin{abstract}
Open-vocabulary object detection aims to detect arbitrary classes via text prompts. Methods without cross-modal fusion layers (non-fusion) offer faster inference by treating recognition as a retrieval problem, \ie, matching regions to text queries in a shared embedding space. In this work, we fully explore this retrieval philosophy and demonstrate its unique advantages in efficiency and versatility through a model family named WeDetect: (1) \textbf{State-of-the-art performance.} \textbf{WeDetect} is a real-time detector with a dual-tower architecture. We show that, with well-curated data and full training, the non-fusion WeDetect surpasses other fusion models and establishes a strong open-vocabulary foundation. (2) \textbf{Fast backtrack of historical data.} \textbf{WeDetect-Uni} is a universal proposal generator based on WeDetect. We freeze the entire detector and only finetune an objectness prompt to retrieve generic object proposals across categories. Importantly, the proposal embeddings are class-specific and enable a new application, \textbf{object retrieval}, supporting retrieval objects in historical data. (3) \textbf{Integration with LMMs for referring expression comprehension (REC).} We further propose \textbf{WeDetect-Ref}, an LMM-based object classifier to handle complex referring expressions, which retrieves target objects from the proposal list extracted by WeDetect-Uni. It discards next-token prediction and classifies objects in a single forward pass. Together, the WeDetect family unifies detection, proposal generation, object retrieval, and REC under a coherent retrieval framework, achieving state-of-the-art performance across 15 benchmarks with high inference efficiency. Code is available at \url{https://github.com/WeChatCV/WeDetect}.
\end{abstract}

\section{Introduction}

Recognition is a central problem in computer vision. At the image level, the field has progressed from closed-set image classification~\cite{resnet, dosovitskiy2020image, jiao2023dilateformer} to open-vocabulary image retrieval powered by large-scale image–text contrastive learning~\cite{clip}. In parallel, open-vocabulary object detection~\cite{GLIP, grounding_dino, fu2025HD-OVD, fu2025llmdet} extends beyond the fixed label spaces of closed-set detectors~\cite{ren2016faster, detr, fu2023asag, fu2024frozen-detr}, allowing recognition and localization of arbitrary categories specified by textual prompts. By aligning region features with text embeddings, open-vocabulary detectors can achieve zero-shot region recognition without task-specific training.

To improve vision–language alignment, recent open-vocabulary object detectors~\cite{GLIP, grounding_dino, fu2025llmdet} employ various deep cross-modal fusion mechanisms. While achieving high accuracy, their computationally intensive fusion layers substantially degrade inference efficiency. Moreover, the fusion makes visual features query-specific, preventing feature sharing across different textual prompts. For example, evaluating Grounding-DINO~\cite{grounding_dino} on LVIS~\cite{gupta2019lvis} containing 1,203 categories with a chunk size of 40 requires 31 separate forward passes, resulting in several seconds of latency per image and limiting practical deployment. 

In contrast, non-fusion methods adopt a dual-tower architecture and enjoy a fast inference speed. We notice a key characteristic of the non-fusion paradigm, i.e., its recognition is similar to the retrieval problem, which matches image regions against text queries in a shared embedding space. This formulation enjoys unique advantages in efficiency and versatility. Motivated by this insight, we fully explore the retrieval-inspired philosophy through a model family named WeDetect: (1) \textbf{WeDetect}, a strong detection foundation with real-time latency and superior open-vocabulary object detection performance; (2) \textbf{WeDetect-Uni}, a universal proposal generator supporting fast backtrack of historical data; and (3) \textbf{WeDetect-Ref}, an LLM-based REC model for complex expression detection. They are demonstrated as follows:

\textbf{WeDetect} is finetuned from a pretrained CLIP model, comprising a text encoder for encoding class names and a visual encoder for extracting multi-scale visual features. Classification is performed via dot products between class embeddings and image grid features. Rather than relying on deep fusion, we employ three key techniques to achieve superior open-vocabulary object detection performance with efficient inference: (1) Model pretraining: WeDetect is finetuned from a strong, well-pretrained CLIP model to inherit robust open-vocabulary capabilities; (2) Model architecture: As mainstream CLIP variants typically adopt a ViT encoder~\cite{dosovitskiy2020image}, whose plain design is suboptimal for detection, we pretrain a CLIP variant with a ConvNeXt~\cite{liu2022convnext} backbone that naturally provides multi-scale features; and (3) Training data: We develop a data engine to curate a high-quality dataset characterized by balanced concepts, exhaustive annotations, and multi-granularity labels, comprising 15M images and 330M bounding boxes. Together, these design choices enable WeDetect to deliver strong open-vocabulary object detection performance and fast inference without resorting to deep fusion.

Building on WeDetect, we introduce a universal proposal generator, \textbf{WeDetect-Uni}. We freeze the entire detector and train only an objectness embedding for classification. As the detector is frozen, the box embeddings corresponding to the top-scoring proposals remain class-specific and can therefore be used for class-specific classification. By caching proposals together with their box embeddings, we enable fast object retrieval via a CLIP-style dot product. On this basis, we propose a new application, \textbf{object retrieval}, which aims to retrieve images containing user-specified objects even when they are small (\eg cigarette butts). This fine-grained, local retrieval task complements CLIP’s conventional image-level retrieval.

To further handle complex expressions in Referring Expression Comprehension (REC), we employ an LMM-based classification model, \textbf{WeDetect-Ref}. Given the top-scoring proposals produced by WeDetect-Uni together with the query expression, WeDetect-Ref retrieves the target objects by applying a newly designed binary classification head over the candidate proposals. The model processes all objects in parallel and the prediction is conducted in a single forward pass, eliminating the time-consuming next-token prediction which decodes objects sequentially. This retrieval-based paradigm avoids the bounding box regression drawback derived from language modeling and the slow inference speed derived from next-token prediction, while fully leveraging LLM’s language understanding and open-vocabulary capabilities to achieve fast and accurate classification.

Leveraging the retrieval-based methodology, the WeDetect family demonstrates strong open-vocabulary capability with exceptionally high inference throughput. Specifically, WeDetect-Tiny attains 37.4 AP on LVIS minival and 31.4 AP on LVIS at 62.5 fps, surpassing YOLO-World-L~\cite{yolo_world} by 2.0 and 4.6 AP, respectively, while YOLO-World-L runs at 54.6 fps. By scaling up the model, WeDetect-Large achieves 49.4 AP on LVIS, outperforming LLMDet~\cite{fu2025llmdet} by 7.4 AP. In the object retrieval task, WeDetect-Uni outperforms CLIP~\cite{clip} by 37.2 F1 scores, showing its unique advantage in fine-grained perception. Moreover, WeDetect-Ref 4B gets an average score of 93.2 on refcoco/+/g~\cite{refcoco}, exceeding Qwen3-VL~\cite{Qwen3-VL} 4B by 6.5 points with a 13× speedup. We hope that this retrieval paradigm can be broadly adopted by the research community.

\section{Related Work}
\subsection{Open-Vocabulary Object Detection}

Open-vocabulary object detection aims to detect objects with text prompts, requiring fine-grained vision-language alignment. To construct a unified vision-language space, previous works mainly focus on four aspects: (1) Training data constructions: GLIP~\cite{GLIP} first unifies object detection and phrase grounding through region-word contrastive pre-training, which can leverage massive image-text pairs for training. The vast vocabulary existing in the web-scale image-text pairs builds a robust vision-language space. Further, constructing hard negative samples~\cite{yao2022detclip, yao2023detclipv2, li2024desco, zhao2024generating} can provide richer supervision and achieve fine-grained alignment. By scaling up data and computation~\cite{yao2023detclipv2, OWL-ST}, models can achieve impressive zero-shot performance. (2) Training objective: In addition to region-word contrastive learning, unifying other language tasks, including mask language modeling~\cite{zhang2022glipv2}, dense captioning~\cite{long2023capdet, yao2024detclipv3}, and co-training with a large language model~\cite{fu2025llmdet}, enriches visual representations with language knowledge, thus creating a stronger open-vocabulary detector. (3) Vision-language fusion layers: Deep vision-language fusion layers~\cite{GLIP, fiber, grounding_dino, ov-dino, prompt-dino} that jointly integrate visual and textual features can further improve vision-language alignment. However, these computationally intensive fusion layers greatly reduce inference efficiency. And the extracted vision features can not be shared across different queries. (4) Model distillation: Other methods~\cite{vild, oadp, fu2025HD-OVD, fu2024frozen-detr} aim to distill the open-vocabulary knowledge from other foundation models. In this work, we revisit the deep fusion architecture and propose a family of models with plain architecture following the retrieval methodology.

\begin{figure*}[t]
  \centering
  \includegraphics[width=0.9\linewidth]{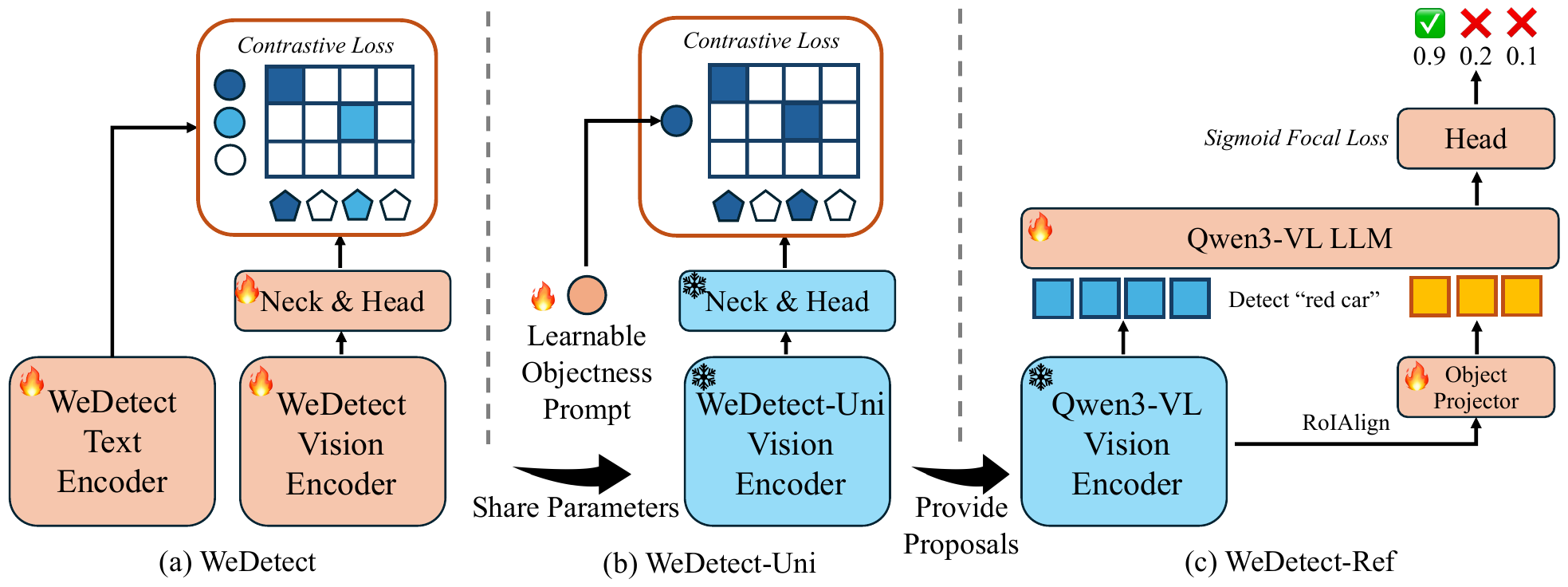}
  \caption{The WeDetect model family: (a) WeDetect is an open-vocabulary object detector with a dual-tower architecture without any multi-modal fusion layers. (b) WeDetect-Uni is a universal proposal generator whose parameters are shared with WeDetect except for a learnable objectness prompt for classification. (c) WeDetect-Ref is an LLM-Based REC model, which can retrieve target objects from proposals provided by WeDetect-Uni corresponding to user-provided expressions.}
  \label{fig:model}
\end{figure*}

\subsection{Large Vision-Language Model}

Large Vision-Language Models (LVLMs)~\cite{bai2025qwen25, Qwen3-VL, chen2024expanding, wang2025internvl35}, pretrained on massive corpus, show not only professional world knowledge and reasoning ability but also superior visual perception and understanding ability. Therefore, LVLMs excel at open-vocabulary perception. To extend LVLMs with region perception ability~\cite{yuan2024osprey, yuan2025videorefer, dam, zhang2024gpt4roi}, each object will be encoded into special tokens separately. However, the language modeling mechanism constrains them for precise object localization, as digits are encoded into separate discrete tokens and are optimized via cross-entropy loss. To eliminate the regression drawback, some methods~\cite{lai2024lisa, tang2025ufo, su2025patch} utilize an extra decoder to decode object tokens, while others~\cite{zhao2024octopus, jiang2024chatrex, jiang2025referring, liu2025vlm} pre-extract some proposals for LLMs to refer to. However, these methods still follow the next-token prediction mechanism, in which objects are decoded sequentially. Therefore, the inference speed is greatly constrained. In this work, we follow the retrieval methodology and utilize the LLM as a classifier to process objects in parallel.

\section{WeDetect: A Strong Detection Foundation}

In this work, we aim to develop a simple and fast open-vocabulary object detector with diverse usages following the retrieval methodology. Based on the goal, we discard the time-consuming fusion layers. In contrast, we adopt the dual-tower architecture from CLIP~\cite{clip} and extend it to region-wise perception. To achieve fine-grained vision-language alignment, we make great efforts in dataset construction and model training, which are detailed as follows.

\subsection{Model Architecture}

WeDetect is a dual-tower architecture model, as shown in \Cref{fig:model}(a). The language encoder is initialized from XLM-RoBERTa~\cite{conneau2020unsupervised} while the vision encoder follows a YOLO-like architecture containing a ConvNeXt~\cite{liu2022convnext} backbone to produce multi-scale features, a CSPRepBiFPAN neck~\cite{li2023yolov6v3}, and a YOLO-World~\cite{yolo_world} contrastive head. The loss functions and label assignment strategy are all the same as YOLO-World, which is a region-text contrastive loss for classification along with a box regression loss. Different from YOLO-World, we do not use any fusion layers within the neck. And classification is simply conducted via the dot product between image grid features and class text embeddings. This simple and elegant architecture ensures a high inference speed.

\begin{figure*}[t]
  \centering
  \includegraphics[width=1\linewidth]{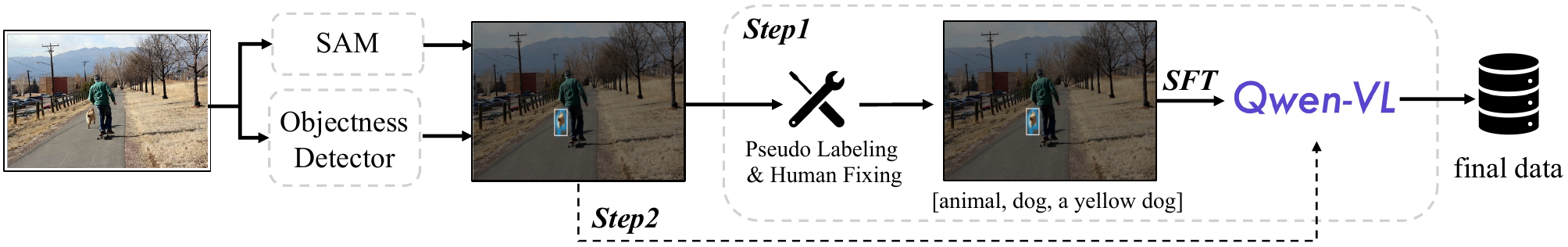}
  \caption{The proposed data engine. We first use an objectness detector to detect all regions of interest along with masks produced by SAM. Then, an LMM is used to generate multi-granularity and instance-specific labels for each object. The LMM is finetuned by us to ensure high-quality labeling and structural output.}
  \label{fig:data_engine}
\end{figure*}

\subsection{Dataset Construction}

A high-quality dataset should be rich in diversity and accurate in annotations. Although open-sourced grounding datasets (\eg GoldG~\cite{GLIP}) contain diverse text expressions, they are limited in dataset size, image diversity, annotation integrity, and annotation diversity. Therefore, we collect a large-scale grounding dataset with balanced-sampled images and well-annotated labels.

\vspace{0.2em}\noindent{\textbf{Source image sampling.}} We first sample source images from various datasets, including SAM-1B~\cite{sam}, LAION~\cite{schuhmann2021laion}, CC12M~\cite{cc12m}, Zero~\cite{xie2023ccmb}, and self-crawled images from licensed websites. The raw captions paired with images (if they exist) are used for selecting some rare nouns to balance the concepts. Totally, source images comprise 15M samples with a wide span of concepts from various domains, ensuring high image diversity.

\vspace{0.2em}\noindent{\textbf{Box annotation pipeline.}} We further propose an automatic data engine to annotate images with high-quality and multi-granularity labels, as shown in \Cref{fig:data_engine}. To ensure high text diversity, we resort to generative methods to generate instance-specific annotations. Specifically, we first train an objectness detector with all available object detection datasets. The objectness detector recalls all objects within the image, which ensures the integrity of annotations. Then, a modern MLLM Qwen2.5-VL 7B~\cite{bai2025qwen25} is used to generate instance-specific hierarchical labels. For example, as shown in \Cref{fig:data_engine}, a dog will be annotated as ``animal, dog, a yellow dog''. In our experiments, these multi-granularity labels greatly enrich the text diversity and improve the performance. To ensure label quality, we finetune the Qwen2.5-VL model with a human-annotated instruction dataset to enhance two crucial abilities: structural output and rejecting recognition. Structural output requires the model to output the label with a fixed template, first outputting coarse-grained labels and then fine-grained labels. And the model with rejecting recognition will not generate labels for erroneous boxes, which also serves as a validation for the previous proposal generation. These boxes will be discarded. Further, to enhance Qwen2.5-VL's local awareness, we highlight the object boundaries in the original image with the mask produced by SAM~\cite{sam} along with the textual box coordinates as the model inputs. Once the annotator is trained, it can annotate remaining images without human supervision.

In summary, our self-annotated dataset contains 15M samples and 330M bounding boxes. We also include other open-sourced object detection and grounding datasets for training, including OpenImagesV6~\cite{kuznetsova2020open}, Objects365 V2~\cite{shao2019objects365}, V3Det~\cite{wang2023v3det}, ImageNetBox~\cite{deng2009imagenet}, and GoldG~\cite{GLIP}. Details are shown in the Appendix.

\subsection{Model Training}

\noindent{\textbf{Staged-wise training method.}} To equip the model with the basic open-vocabulary ability, we first pretrain the model with a CLIP-like image-level contrastive objectiveness on a large-scale image-text dataset. The resulting checkpoints are used for initializing the visual backbone and the language encoder of WeDetect. As the neck and the head are still randomly initialized, in the second stage, we freeze the visual backbone and the language encoder, and only train the remaining components. In the last stage, all parameters are trained in an end-to-end manner. This staged-wise training method can fully leverage the pretraining knowledge while adopting it from image-level to region-level.

\vspace{0.2em}\noindent{\textbf{Multi-granularity label sampling.}} In our self-collected dataset, each object is annotated with multi-granularity labels. We propose a multi-granularity label sampling method as a kind of data augmentation, which independently samples a label from the candidate list for each object during each training iteration. The fine-grained and diverse text labels will not only provide rich supervision for the single object but also construct a diverse and training-time-specific vocabulary for the whole batch, providing diverse negative samples for learning. This data augmentation greatly boosts the open-vocabulary performance.

\section{WeDetect-Uni: A Universal Proposal Generator}

\noindent{\textbf{Extracting arbitrary objects via a universal objectness prompt.}} In this section, we extend WeDetect to a universal proposal generator, WeDetect-Uni, without user-provided text prompts. Specifically, as shown in \Cref{fig:model}(b), we freeze the entire detector and train a universal objectness prompt for classification, which is kind of linear probing finetuning. Based on WeDetect's discriminative features, only a single learnable prompt is needed for high recall rates. Importantly, different from other class-agnostic proposal networks~\cite{ren2016faster}, the box embeddings corresponding to the top-scoring proposals are still class-specific, which can be used for classification and serve as the basis for the following new application.

\begin{figure}[t]
  \centering
  \includegraphics[width=1\linewidth]{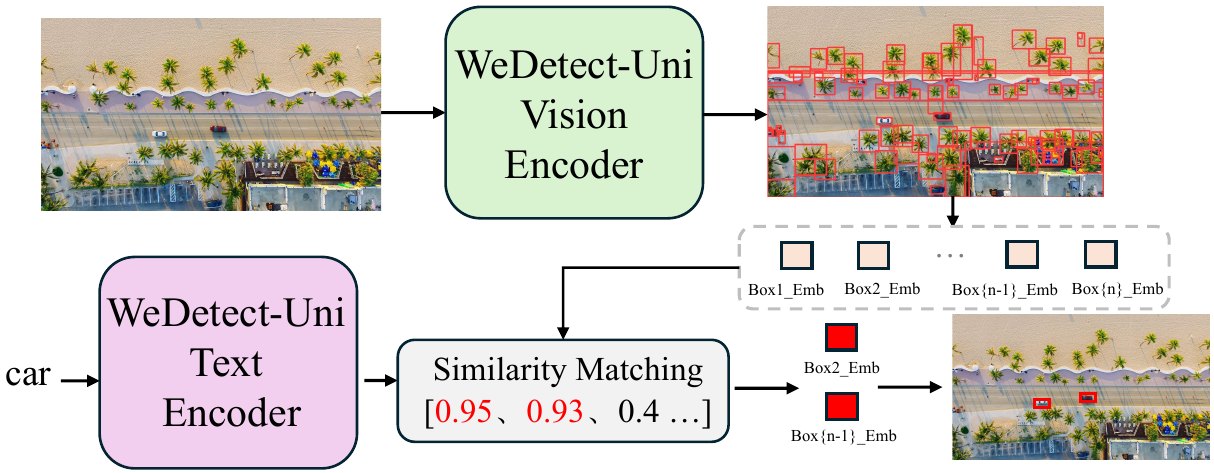}
  \caption{Illustration of applying WeDetect-Uni to the object retrieval task. We first use WeDetect-Uni to extract some regions of interest. And the box embeddings corresponding to the top-scoring proposals are cached to represent the image. Once a query comes, only a simple dot production is needed for fast retrieval.}
  \label{fig:retrieval}
\end{figure}

\vspace{0.2em}\noindent{\textbf{A new local object retrieval task.}} In this work, we propose a new task, object retrieval, in which models should retrieve all images containing a user-specified object category from a database. Different from the image-text retrieval task, where the query focuses on the global image semantics, the object retrieval task pays attention to the local semantics, such as small objects like ``cigarette butts'', which complements the image-level retrieval task. This task is valuable for keyword image retrieval, image content examination and verification, and other real-world applications. For evaluation, the new task can be conducted on common object detection datasets, where class names are user queries, the whole validation set is the database, and images containing class-specific annotations constitute the ground truth image set. The evaluation is conducted between the ground truth image set and the predicted image set, and the metrics are precision, recall, and F1 score.

\vspace{0.2em}\noindent{\textbf{Applying WeDetect-Uni to the object retrieval task.}} Different from CLIP, where each image is encoded as a single embedding, we use a set of object embeddings to represent an image. As shown in \Cref{fig:retrieval}, we use WeDetect-Uni to detect all regions of interest. The box embeddings corresponding to the top-scoring proposals are pre-extracted and cached to represent the image. Once a new query arrives, a simple dot production is needed for fast retrieval.

\section{WeDetect-Ref: An LLM-Based REC Model}

\begin{table*}[t]
  \centering
  \vspace{-0.6em}
  \caption{Zero-shot detection performance. Metrics on LVIS val~\cite{gupta2019lvis} and minival~\cite{kamath2021mdetr} are fixed AP~\cite{dave2021evaluating}. WeDetect achieves state-of-the-art performance across various model scales. \textcolor{gray}{Gray} numbers indicate including COCO data in training. FPS is tested on COCO dataset.}
  \vspace{-0.6em}
  \resizebox{\linewidth}{!}{
      \begin{tabular}{llccc|cccc|cccc|c|c|c|c}
        \hline
        \multirow{2}{*}{Method} & \multirow{2}{*}{Backbone} & \multirow{2}{*}{Resolution} & \multirow{2}{*}{\#Params} & \multirow{2}{*}{FPS} & \multicolumn{4}{c|}{LVIS$^{\text{minival}}$}  & \multicolumn{4}{c|}{LVIS}  & COCO & COCO-O & ODinW13 & ODinW35 \\
        & & & & & AP & AP$_{r}$ & AP$_{c}$ & AP$_{f}$ & AP & AP$_{r}$ & AP$_{c}$ & AP$_{f}$ & AP & AP & AP & AP \\
        \hline
        YOLO-World-L~\cite{yolo_world} & YOLOv8-L & 640*640 & 48M & 54.6& 35.4 & 27.6 & 34.1 & 38.0  & 26.8  & 19.8 & 23.6 & 33.4 & \textbf{44.9} & 32.5 & 38.4 & 17.1 \\
        YOLOE-8-L~\cite{wang2025yoloe} & YOLOv8-L & 640*640 & 45M & - & 35.9 & 33.2 & 34.8 & 37.3 & - & - & - & - & - & - & - & - \\
        \rowcolor{ours} WeDetect-Tiny & ConvNext-T & 640*640 & 33M & 62.5 & \textbf{37.4} & \textbf{33.3}& \textbf{36.8}& \textbf{38.8}& \textbf{31.4} &\textbf{24.7} &\textbf{29.2} & \textbf{36.8}& \textbf{44.9} & \textbf{38.6} & \textbf{46.4} & \textbf{21.1} \\
        \hline
        GLIP~\cite{GLIP} & Swin-T & 800*1333 & 232M & 5.4 & 26.0 & 20.8 & 21.4 & 31.0 & 17.2 & 10.1 & 12.5 & 25.2 & 46.1 & 29.0 & 46.5 & 19.6 \\
        Grounding-DINO~\cite{grounding_dino} & Swin-T & 800*1333 & 172M & 6.0 & 27.4 & 18.1 & 23.3 & 32.7 & 20.1 & 10.1 & 15.3 & 29.9 & 48.4 & 37.6 & 51.4 & 22.7 \\
        DetCLIP~\cite{yao2022detclip} & Swin-T & 800*1333 & - & - & 35.9 & 33.2 & 35.7 & 36.4 & 28.4 & 25.0 & 27.0 & 28.4 & - & - & 43.3 & - \\
        DetCLIPv2~\cite{yao2023detclipv2} & Swin-T & 800*1333 & - & - & 40.4 & 36.0 & 41.7 & 40.4 & 32.8 & 31.0 & 31.7 & 34.8 & - & - & - & - \\
        DetCLIPv3~\cite{yao2024detclipv3} & Swin-T & 800*1333 & - & - & 47.0 & \textbf{45.1} & \textbf{47.7} & 46.7 & 38.9 & 37.2 & 37.5 & 41.2 & 47.2 & 38.5 & - & - \\
        T-Rex2~\cite{T-Rex2} & Swin-T & 800*1333 & - & - & 42.8 & 37.4 & 39.7 & 46.5 & 34.8 & 29.0 & 31.5 & 41.2 & 45.8 & - & - & 18.0 \\
        OV-DINO~\cite{ov-dino} & Swin-T & 800*1333 & - & - & 40.1 & 34.5 & 39.5 & 41.5 & 32.9 & 29.1 & 30.4 & 37.4 & 50.2 & - & - & - \\
        MM-GDINO~\cite{mm_GDINO} & Swin-T & 800*1333 & 172M & 6.0 & 41.4 & 34.2 & 37.4 & 46.2 & 31.9 & 23.6 & 27.6 & 40.5 & 50.4 & 34.0 & 52.5 & 23.1 \\
        LLMDet~\cite{fu2025llmdet} & Swin-T & 800*1333 & 172M & 6.0 & 44.7 & 37.3 & 39.5 & \textbf{50.7} & 34.9 & 26.0 & 30.1 & 44.3 & \textcolor{gray}{55.6} & 36.1 & 52.1 & 23.8  \\
        DINO-X Edge~\cite{ren2024dinox} & EfficientViT-L2 & 640*640 & - & 19.8 & 44.5 & 41.4 & 47.3 & 42.6 & 38.4 & \textbf{38.9} & 38.3 & 38.2 & 48.7 & - & - & - \\
        \rowcolor{ours} WeDetect-Base & ConvNext-B & 640*640 & 176M & 35.1& \textbf{47.3} &43.5 &45.9 & 49.3& \textbf{41.4} & 35.2& \textbf{39.5}& \textbf{46.2}& \textbf{52.1} & \textbf{44.1} & \textbf{53.1} & \textbf{24.6} \\
        \hline
        GLIP~\cite{GLIP} & Swin-L & 800*1333 & 430M & 3.1 & 37.3 & 28.2 & 34.3 & 41.5 & 26.9 & 17.1 & 23.3 & 36.4 & 49.8 & - & - & - \\
        Grounding-DINO~\cite{grounding_dino} & Swin-L & 800*1333 & 343M & 2.1 & 33.9 & 22.2 & 30.7 & 38.8 & – & – & – & – & 52.5 & - & - & \textbf{26.1} \\
        DetCLIP~\cite{yao2022detclip} & Swin-L & 800*1333 & - & - & 38.6 & 36.0 & 38.3 & 39.3 & 28.4 & 25.0 & 27.0 & 31.6 & - & - & 50.0 & 24.9 \\
        DetCLIPv2~\cite{yao2023detclipv2} & Swin-L & 800*1333 & - & - & 44.7 & 43.1 & 46.3 & 43.7 & 36.6 & 33.3 & 36.2 & 38.5 & - & - & - & - \\
        DetCLIPv3~\cite{yao2024detclipv3} & Swin-L & 800*1333 & - & - & 48.8 & 49.9 & 49.7 & 47.8 & 41.4 & 41.4 & 40.5 & 42.3 & 48.5 & 48.8 & - & - \\
        LLMDet~\cite{fu2025llmdet} & Swin-L & 1000*1560 & 343M & 2.1 & 50.6 & 41.7 & 46.2 & \textbf{56.1} & 42.0 & 31.6 & 38.8 & 50.2 & \textcolor{gray}{59.2} & \textbf{53.2} & 53.3 & 24.6 \\
        T-Rex2~\cite{T-Rex2} & Swin-L & 800*1333 & - & - & 54.9 & 49.2 & \textbf{54.8} & \textbf{56.1} & 45.8 & 42.7 & 43.2 & 50.2 & 52.2 & - & 50.3 & 22.0 \\
        \rowcolor{ours} WeDetect-Large & ConvNext-L & 1280*1280 & 490M & 6.0 & \textbf{55.0} &\textbf{51.1} & 54.5& \textbf{56.1}& \textbf{49.4} & \textbf{43.3} & \textbf{48.2}&\textbf{53.5} & \textbf{54.5} & 47.0 & \textbf{53.4} & 25.8 \\
        \hline
      \end{tabular}
    }
  \vspace{-0.6em}
  \label{tab:lvis_result}
\end{table*}

\subsection{Formulating REC as Retrieval}

In real-world applications, user queries can be much more complex by specifying appearances, materials, locations, and even requiring some reasoning ability based on common sense, which needs deep language and semantic understanding and poses great challenges to traditional detectors. Inspired by recent frontier large vision-language models, we aim to use them to handle the complex referring expression comprehension (REC) task. However, two crucial challenges should be tackled: First, as large language models (LLMs) are trained with the language modeling objective rather than the regression objective, in which digits are represented as discrete tokens and optimized by cross-entropy loss, making the model less sensitive to the precise bounding boxes. Second, LLMs work in a next-token prediction manner, in which tokens should be generated sequentially with multiple model forward passes, resulting in extremely long model latency. Motivated by VideoITG~\cite{wang2025videoitg}, we formulate the REC task as a retrieval task and simply use the large language models as a classifier to retrieve target objects from a pre-extracted candidate list.

Specifically, we first use WeDetect-Uni to extract some objects of interest as the object candidate list $\{B_i\}_{i=1}^n$. As shown in \Cref{fig:model}(c), for each object, we extract its multiscale RoI features from the MLLM's visual encoder and then compress them to a single token $\{o_i\}_{i=1}^n$ via a linear object projector. The full image tokens $I$, the user query $q$, and object tokens $\{o_i\}_{i=1}^n$ are concatenated and sent to the LLM for classification. The classification is conducted by applying a newly introduced binary classification head over the hidden embeddings of object tokens to decide whether the object belongs to the query:
\begin{align}
    \{h_i\}_{i=1}^n &= \text{LLM}(I, q, \{o_i\}_{i=1}^n), \\
    \{s_i\}_{i=1}^n = \text{Sigmoid}&(\text{Classifier}(\{h_i\}_{i=1}^n)) \in [0, 1],
\end{align}
where $\{s_i\}_{i=1}^n$ are classification scores for each object.

In this paradigm, the LLM only acts as a classifier to retrieve target objects corresponding to the query from a class agnostic candidate list, which enjoys two advantages: First, this retrieval-based paradigm avoids the bounding box regression drawback derived from language modeling while fully leveraging LLM’s language understanding and open-vocabulary capabilities to achieve fast and accurate classification. Second, this retrieval-based paradigm discards the next-token prediction mechanism so that the prediction can be conducted in a single model forward pass, achieving a superior inference speed. Although the retrieval-based paradigm is briefly explored by previous works~\cite{zhao2024octopus, jiang2024chatrex}, they still follow the next-token prediction mechanism thus limiting the inference speed.

\subsection{A Three-Stage Training Recipe}

In this work, we use Qwen3-VL~\cite{Qwen3-VL} as our base MLLM and extend it with fine-grained region perception ability by a three-stage training recipe.

\vspace{0.2em}\noindent{\textbf{Stage 1: Region projector training.}} In our model, we introduce a new special token ``$\langle$object$\rangle$'' as a placeholder for each object. For each object, we use RoIAlign to extract its multi-scale features from the last visual feature map and other deep stack visual feature maps. We use a linear layer as the region projector to compress RoI features into a single token. The bounding boxes will be encoded as position embeddings and added to object tokens. Then, the placeholder is replaced by the object token before sending to the LLM. In this stage, we only finetune the newly introduced region projector with a 700K image-level and region-level caption dataset. The data format is shown as follows:

\noindent\framebox[\linewidth]{%
  \parbox{0.9\linewidth}{
    user: $\langle$image$\rangle$ Describe the object $\langle$object$\rangle$ briefly. \\
    assistant: far right guy.
  }%
}

\vspace{0.2em}\noindent{\textbf{Stage 2: Region perception finetuning.}} In this stage, we further finetune the LLM and projector to better align with the object tokens, while the vision encoder is still frozen. In addition to caption data, we include many other image-level and region-level instruction tuning data (around 1.7M data) with the same format as above. After finetuning, the LLM can perceive specific objects accurately.

\vspace{0.2em}\noindent{\textbf{Stage 3: Region classification finetuning.}} In the last stage, we finetune the LLM into a classifier model. We discard the original language modeling head, and instead train a binary classification head. The classification head is only applied over the hidden embeddings of object tokens. To process a list of objects at once, we formulate the data template as follows:

\noindent\framebox[\linewidth]{%
  \parbox{0.9\linewidth}{
    user: $\langle$image$\rangle$ Please detect the ``CLASSNAME'' in the image. \\
    assistant: $\langle$object$\rangle$$\langle$object$\rangle$$\langle$object$\rangle$$\langle$object$\rangle$$\langle$object$\rangle$
  }%
}
where the ``CLASSNAME'' will be replaced by the user-provided categories or expressions and the number of $\langle$object$\rangle$ equals the number of proposals. We collect some open-sourced object detection datasets and referring expression comprehension datasets containing 4M samples for training. Details are summarized in the Appendix. For the loss function, we use sigmoid focal loss~\cite{ross2017focal} with IoU as soft labels. All proposals with an IoU greater than 0.5 with any ground truth bounding box are treated as positive samples. In this stage, the trainable parameters are also the LLM and the projector. Other implementation details will be provided in the Appendix.

\section{Experiment}

\begin{table*}[t]
  \centering
  \caption{Evaluation results on common referring expression comprehension datasets. The evaluation metric for RefCOCO, RefCOCO+, and  RefCOCOg is the Top-1 accuracy. FPS is tested on the RefCOCO dataset.}
  \vspace{-0.6em}
  \resizebox{0.97\linewidth}{!}{
      \begin{tabular}{l|c|ccccccccc|cccc}
        \hline
        \multirow{2}{*}{Method} & \multirow{2}{*}{FPS}  & \multicolumn{3}{c}{RefCOCO} & \multicolumn{3}{c}{RefCOCO+} & \multicolumn{2}{c}{RefCOCOg} &  & \multicolumn{4}{c}{HumanRef} \\
        & & val & testA & testB & val & testA & testB & val & test & Avg. & P & R & DF1 & Rej. \\
        \hline
        Grounding-DINO-L~\cite{grounding_dino} & 3.1 & 90.6 & 93.2 & 88.2 & 82.8 & 89.0 & 75.9 & 86.1 & 87.0 & 86.6 & 33.1 & 75.2 & 23.3 & - \\
        \hline
        Qwen2.5-VL 3B~\cite{bai2025qwen25} & - & 89.1 & 91.7 & 84.0 & 82.4 & 88.0 & 74.1 & 85.2 & 85.7 & 85.0 & - & - & - & - \\
        Qwen2.5-VL 7B~\cite{bai2025qwen25} & - & 90.0 & 92.5 & 85.4 & 84.2 & 89.1 & 76.9 & 87.2 & 87.2 & 86.6 & 68.5 & 52.5 & 56.2 & 7.1 \\
        InternVL2.5-8B~\cite{chen2024expanding} & - & 90.3 & 94.5 & 85.9 & 85.2 & 91.5 & 78.8 & 86.7 & 87.6 & 87.6 & 37.9 & 29.8 & 31.9 & 54.9 \\
        InternVL3.5-8B~\cite{wang2025internvl35}  & - & 92.4 & 94.7 & 88.7 & 87.9 & 92.4 & 82.4 & 89.6 & 89.4 & 89.7 & - & - & - & - \\
        InternVL3.5-38B~\cite{wang2025internvl35} & - & 90.3 & 91.8 & 89.0 & 87.5 & 90.0 & 84.7 & 89.7 & 89.9 & 89.1 & - & - & - & - \\
        InternVL3.5-241B-A28B~\cite{wang2025internvl35} & - & 94.1 & 96.3 & 91.5 & \textbf{91.6} & 94.6 & \textbf{86.9} & 92.0 & 92.1 & 92.4 & - & - & - & - \\
        Qwen3-VL-235B-A22B Thinking~\cite{Qwen3-VL}  & - & - & - & - & - & - & - & - & - & 92.4 & - & - & - & - \\
        \hline
        Octopus 7B~\cite{zhao2024octopus} & - & 89.0 & 92.6 & 83.4 & 83.6 & 89.4 & 76.0 & 84.3 & 86.3 & 85.6 & - & - & - & - \\
        VLM-R1 3B~\cite{shen2025vlmr1} & - & 90.1 & 92.3 & 85.2 & 84.2 & 89.4 & 76.8 & 85.6 & 86.8 & 86.3 & - & - & - & - \\
        Rex-Omni 3B~\cite{jiang2025detect} & - & 86.6 & 89.5 & 82.8 & 79.6 & 84.8 & 71.4 & 85.3 & 86.2 & 83.3 & 79.3 & 80.1 & 75.6 & - \\
        ChatRex 7B~\cite{jiang2024chatrex} & - & 91.0 & 94.1 & 87.0 & 89.8 & 91.9 & 79.3 & 89.8 & 90.0 & 89.1 & 72.2 & 50.4 & 55.6 & 0.0 \\
        VLM-FO1 3B~\cite{liu2025vlm} & - & 91.1 & 93.7 & 87.6 & 86.4 & 91.9 & 80.6 & 88.9 & 88.3 & 88.6 & \textbf{87.1} & 83.3 & \textbf{82.6} & - \\
        RexSeek 7B~\cite{jiang2025referring} & - & - & - & - & - & - & - & 84.0 & 84.4 &  - & 85.8 & 85.9 & 82.4 & 54.1 \\
        \hline
        Qwen3-VL 2B~\cite{Qwen3-VL} & 0.6 & 88.2 & 91.0 & 83.1 & 78.6 & 85.2 & 70.4 & 84.7 & 85.0 & 83.3 & 69.7 & 58.2 & 60.2 & 20.6 \\
        Qwen3-VL 4B~\cite{Qwen3-VL} & 0.4 & 90.7 & 92.2 & 86.7 & 82.9 & 89.4 & 75.6 & 87.3 & 87.7 & 86.6 & 76.7 & 65.9 & 67.8 & 39.1 \\
        \rowcolor{ours} WeDetect-Ref 2B & 6.6 & 94.3 & 95.6 & 92.6 & 88.1 & 92.6 & 83.1 & 92.0 & 92.2 & 91.3 & 84.7 & 85.1 & 79.8 & 61.0 \\
        \rowcolor{ours} WeDetect-Ref 4B & 5.3 & \textbf{95.6} & \textbf{96.7} & \textbf{93.6} & 90.5 & \textbf{94.8} & 86.8 & \textbf{93.8} & \textbf{93.9} & \textbf{93.2} & 86.3 & \textbf{87.1} & 81.8 & \textbf{64.1} \\
        \hline
      \end{tabular}
    }
  \vspace{-0.6em}
  \label{tab:ref}
\end{table*}

\begin{table}[t]
  \centering
  \caption{Zero-shot recall rates on object detection datasets. \textcolor{gray}{Gray} numbers indicate including the target data in training.}
  \vspace{-0.6em}
  \resizebox{0.95\linewidth}{!}{
      \begin{tabular}{l|cc|cc|cc}
        \hline
        \multirow{2}{*}{Method}  & \multicolumn{2}{c|}{COCO} & \multicolumn{2}{c|}{LVIS}  & \multicolumn{2}{c}{PACO-LVIS}  \\
        & AR$_{50}$ & AR & AR$_{50}$ & AR & AR$_{50}$ & AR \\
        \hline
        \rowcolor{Gray} 100 proposals & & & & & & \\ 
        \hline
        MAVL~\cite{mavl} & 67.3 & 40.4 & 40.7 & 22.3 & 24.5 & 12.5 \\
        OLN~\cite{oln} & 71.9 & 47.5 & 35.8 & 21.4 & 26.7 & 14.8 \\
        RPN-R50~\cite{ren2016faster} & \textcolor{gray}{75.7} & \textcolor{gray}{46.1} & 39.3 & 22.9 & 30.0 & 15.9 \\
        UPN (fine-grained)~\cite{jiang2024chatrex} & 89.6 & 69.2 & 65.0 & 49.0 & 38.3 & 27.2 \\
        UPN (coarse-grained)~\cite{jiang2024chatrex} & \textcolor{gray}{90.6} & \textcolor{gray}{69.7} & \textcolor{gray}{62.0} & \textcolor{gray}{46.6} & 37.8 & 26.6 \\
        \rowcolor{ours} WeDetect-Base-Uni & 87.9 & 66.7 & 67.4 & 50.8 & 37.1 & 25.7 \\
        \rowcolor{ours} WeDetect-Large-Uni & \textbf{89.7} & \textbf{69.3} & \textbf{70.9} & \textbf{56.3} & \textbf{38.4} & \textbf{27.9} \\
        \hline
        \rowcolor{Gray} 300 proposals & & & & & & \\ 
        \hline
        MAVL~\cite{mavl} & 69.7 & 41.2 & 44.1 & 23.5 & 27.9 & 13.4 \\
        OLN~\cite{oln} & 79.5 & 52.4 & 44.3 & 26.0 & 37.0 & 19.6 \\
        RPN-R50~\cite{ren2016faster} & \textcolor{gray}{86.2} & \textcolor{gray}{53.5} & 53.4 & 31.4 & 44.7 & 23.7 \\
        UPN (fine-grained)~\cite{jiang2024chatrex} & 95.0 & 72.9 & 78.9 & 57.2 & \textbf{54.4} & 35.5 \\
        UPN (coarse-grained)~\cite{jiang2024chatrex} & \textcolor{gray}{95.4} & \textcolor{gray}{73.2} & \textcolor{gray}{76.6} & \textcolor{gray}{55.6} & 53.3 & 34.8 \\
        \rowcolor{ours} WeDetect-Base-Uni & 92.3 & 69.6 & 77.9 & 56.9 & 50.2 & 32.1 \\
        \rowcolor{ours} WeDetect-Large-Uni & \textbf{95.3} & \textbf{73.2} & \textbf{84.2} & \textbf{65.1} & 53.3 & \textbf{35.8} \\
        \hline
      \end{tabular}
    }
    \vspace{-0.6em}
  \label{tab:recalls}
\end{table}

\subsection{Main Result}

\noindent{\textbf{WeDetect achieves superior open-vocabulary object detection performance with a faster inference speed.}} To demonstrate the open-vocabulary capacity, we evaluate WeDetect on various object detection benchmarks in a zero-shot manner, including LVIS~\cite{gupta2019lvis}, COCO~\cite{coco}, COCO-O~\cite{mao2023coco}, and ODinW~\cite{li2022elevater}. LVIS is a large vocabulary dataset with 1203 classes and a long-tail distribution, requiring recognition of a wide span of objects. COCO contains 80 common object categories in everyday scenes, while COCO-O retains the same categories but extends them to six distinct domains, posing a challenge for cross-domain generalization. ODinW includes 35 diverse object detection datasets with different vocabularies, offering a comprehensive test of zero-shot transferability. As shown in \Cref{tab:lvis_result}, WeDetect achieves state-of-the-art performance across different model scales. Specifically, WeDetect-Tiny outperforms YOLO-World-L~\cite{yolo_world} by 2.0 AP on LVIS minival, 4.6 AP on LVIS, 6.1 AP on COCO-O, 8.0 AP on ODinW13, and 4.0 AP on ODinW35, while running at a faster speed. When scaling up model size, WeDetect-Large outperforms the previous SOTA model T-Rex2~\cite{T-Rex2} by 3.6 AP on the challenging LVIS benchmark. These results highlight WeDetect’s superior open-vocabulary recognition ability. More importantly, without cross-modal fusion layers, WeDetect runs at an extremely fast speed. WeDetect-Tiny runs at a 62.5 fps, surpassing YOLO-World, despite the latter is optimized for efficient inference. Further, WeDetect-Base and WeDetect-Large exceed GroundingDINO~\cite{grounding_dino} by 6 times and 3 times in speed but with higher performance. These results demonstrate the superior advantages of the dual-tower architecture.

\vspace{0.2em}\noindent{\textbf{WeDetect-Uni gets high recall rates with only a learnable prompt.}} Based on the strong detection foundation WeDetect, WeDetect-Uni only trains a learnable prompt for universal proposal generation. We evaluate the recall rates on three benchmarks: COCO~\cite{coco}, LVIS~\cite{gupta2019lvis}, and PACO-LVIS~\cite{ramanathan2023paco}. Note that the three benchmarks share the same images but with different annotation granularities. COCO has only 80 classes, while LVIS extends the vocabulary to 1203 classes, and PACO further annotates object parts. The multi-granularity annotations construct an ideal benchmark for universal proposal generation. As shown in \Cref{tab:recalls}, WeDetect-Large-Uni achieves the highest recall rates on all datasets with a frozen detector, which demonstrates the highly discriminative features of WeDetect.

\vspace{0.2em}\noindent{\textbf{WeDetect-Uni enjoys unique advantages in the region-wise object retrieval task.}} In this work, we propose a new application, denoted as object retrieval, which aims to retrieve images with user-specified objects. We use common object detection datasets COCO val~\cite{coco} and LVIS val~\cite{gupta2019lvis} as the benchmark datasets and use the category names as queries. The precision, recall, and F1 scores are first computed within each class and then averaged across different classes. As LVIS is a federated dataset where not all classes within the image are annotated, we only compute the recall rates on it. We select OpenAI CLIP ViT-large-patch14-336~\cite{clip}, HQ-CLIP-base~\cite{wei2025hqclip} which includes hard negative samples for training, and FG-CLIP2-so400M~\cite{xie2025fgclip2} which is optimized for fine-grained perception as the image-level perception model for comparisons. As shown in \Cref{tab:object_retrieval}, our WeDetect-Large-Uni with 300 proposals significantly outperforms the image-level baselines, showing that the object retrieval task is complementary to the image-text retrieval task and WeDetect-Uni enjoys unique advantages on fine-grained perception.

\begin{table}[t]
  \centering
  \caption{Zero-shot object retrieval results on common object detection datasets. As LVIS~\cite{gupta2019lvis} is a federated dataset, we only compute the recall rates on it. To prevent setting a low threshold to get a high recall rate, we directly use the threshold used in COCO to evaluate LVIS.}
  \vspace{-0.6em}
  \resizebox{0.97\linewidth}{!}{
      \begin{tabular}{l|c|ccc|c}
        \hline
        \multirow{2}{*}{Method} & \multirow{2}{*}{thre.}  & \multicolumn{3}{c|}{COCO} & \multicolumn{1}{c}{LVIS} \\
        & & P & R & F1 & R \\
        \hline
        OpenAI CLIP~\cite{clip} & 0.550 & 60.0 & 46.4 & 46.4 & 30.4 \\
        HQ-CLIP~\cite{wei2025hqclip} & 0.550 & 59.9 & 59.2 & 52.2 & 41.3  \\
        FG-CLIP2~\cite{xie2025fgclip2} & 0.001 & 67.9 & 62.4 & 57.7 & 43.1  \\
        \rowcolor{ours} WeDetect-Base-Uni & 0.200 & 82.5 & 83.9 & 82.5 & 51.1 \\
        \rowcolor{ours} WeDetect-Large-Uni & 0.200 & \textbf{82.6} & \textbf{85.6} & \textbf{83.6} & \textbf{57.5} \\
        \hline
      \end{tabular}
    }
    \vspace{-0.6em}
  \label{tab:object_retrieval}
\end{table}

\vspace{0.2em}\noindent{\textbf{WeDetect-Ref excels in REC tasks with many fewer parameters and a much faster inference speed.}} We select RefCOCO~\cite{refcoco}, RefCOCO+~\cite{refcoco+}, RefCOCOg~\cite{refcocog}, and HumanRef~\cite{jiang2025referring} as the REC benchmarks. As shown in \Cref{tab:ref}, our WeDetect-Ref 4B achieves the highest 93.2 average scores on refcoco/+/g with only 4B parameters and top 100 proposals from WeDetect-Base-Uni, outperforming our baseline Qwen3-VL 4B~\cite{Qwen3-VL} by 6.6 points and other much larger models with thinking ability. Importantly, since we formulate the REC task as a retrieval task and discard the next-token prediction mechanism, WeDetect-Ref 4B runs at an extremely fast speed, exceeding Qwen3-VL 4B by 13 times and even faster than Grounding-DINO-L~\cite{grounding_dino}. In our scenarios, each image will be represented as 900-1600 tokens and 100 proposals are used, containing around 1000 tokens per image. The plain and simple architecture (containing only self-attention layers) can be easily accelerated by modern GPUs and Flash Attention~\cite{dao2023flashattention2}. Further, the inference time of WeDetect-Ref is consistent with the number of target objects, while the time for methods with the next-token prediction will increase linearly. The advantages in both accuracy and inference speed demonstrate the effectiveness of the new paradigm.

\vspace{0.2em}\noindent{\textbf{WeDetect-Ref also performs well in multi-class multi-instance object detection tasks.}} Although object detection is a relatively easy task for traditional detectors, it poses great challenges to LMMs for two reasons: First, as images may contain multiple instances, LMMs with next-token prediction tend to predict a small part of objects, resulting in low recall rates. Second, recent LMMs tend to predict objects for every query, failing to reject negative queries with no objects existing in the image. Therefore, Qwen2.5-VL 7B~\cite{bai2025qwen25} gets only 17.7 AP on the COCO dataset. In contrast, WeDetect-Ref retrieves objects from a candidate list and each object is decoded independently, which ensures high recall rates and it is the first LMM exceeding 50 AP on the COCO dataset, for the first time matching the performance of traditional object detectors. WeDetect-Ref also achieves a high 47.3 mAP on ODinW13. We provide more implementation details in the Appendix.

\begin{table}[t]
  \centering
  \caption{Evaluating WeDetect-Ref on common object detection datasets. * indicates evaluation under a simplified setting where only ground-truth categories are queried.}
  \vspace{-0.6em}
  \resizebox{\linewidth}{!}{
      \begin{tabular}{l|cccc|c}
        \hline
        \multirow{2}{*}{Method} & \multicolumn{4}{c|}{COCO} & ODinW13 \\
        & AP & AP$_s$ & AP$_m$ & AP$_l$ & AP \\
        \hline
        Grounding-DINO-T~\cite{grounding_dino} & 48.4 & - & - & - & 51.4 \\
        \hline
        Qwen2.5-VL 7B~\cite{bai2025qwen25} & 17.7 & - & - & - & 37.3$^*$ \\
        ChatRex 7B~\cite{jiang2024chatrex} & 48.2$^*$ & - & - & - & - \\
        PaDT Pro 7B~\cite{su2025patch} & 39.0 & - & - & - & - \\
        VLM-FO1 3B~\cite{liu2025vlm} & 44.4 & - & - & - & 44.0 \\
        LMM-Det 7B~\cite{li2025lmm} & 47.5 & 34.7 & 51.8 & 60.3 & - \\
        Qwen3-VL 8B~\cite{Qwen3-VL} & - & - & - & - & 44.7 \\
        \rowcolor{ours} WeDetect-Ref 4B & 50.0 & 34.7 & 57.6 & 69.2 & 47.3 \\
        \hline
      \end{tabular}
    }
  \vspace{-0.6em}
  \label{tab:det}
\end{table}

\subsection{Ablation Study}

\begin{table}[t]
  \centering
  \caption{Ablation studies on WeDetect-Base. Experiments are performed on LVIS minival.}
  \vspace{-0.6em}
  \resizebox{\linewidth}{!}{
      \begin{tabular}{c|l|cccc}
        \hline
        Exp & Model & AP & AP$_{r}$ & AP$_{c}$ & AP$_{f}$ \\
        \hline
        \rowcolor{ours} & WeDetect-Base & 47.3 &43.5 &45.9 & 49.3  \\
        (1) & w/o coarse-grained labels & 46.4 & 41.9 & 44.7 & 48.6  \\
        (2) & w/o fine-grained labels &  45.1 & 41.1 & 43.2  & 47.8  \\
        (3) & w/o staged-wise training & 45.5 & 41.2 & 43.7 & 47.9  \\
        \hline
      \end{tabular}
    }
  \label{tab:ab1}
  \vspace{-0.6em}
\end{table}

\noindent{\textbf{Effect of different training strategies for WeDetect.}} In this work, we initialize WeDetect with a pretrained CLIP model and finetune it with our grounding dataset with multi-granularity labels. For multi-granularity labels, we randomly sample one of the last two labels from the list, the finest one and the second finest one. Without multi-granularity labels, the text diversity is greatly reduced and can not achieve fine-grained vision-language alignment. In \Cref{tab:ab1} (1) and (2), it reduces 1.6 AP$_r$ with only the fine-grained labels and reduces 2.2 AP with only the coarse-grained labels. Using multi-granularity labels simultaneously achieves the highest performance. We also propose a staged-wise training recipe by first training the random-initialized head and neck and then training the model as a whole. In \Cref{tab:ab1} (3), without initializing the head and neck, the pretrained CLIP features will be disturbed, decreasing 1.8 AP.

\vspace{0.2em}\noindent{\textbf{Effect of different design choices for WeDetect-Ref.}} As shown in \Cref{tab:ab4}: (1) Sigmoid focal loss is a standard loss function in object detection and it is also suitable for WeDetect-Ref. Replacing it with binary cross-entropy (BCE) loss leads to a notable drop of 4.4 AP on COCO. (2) Negative supervision is crucial for learning a robust model. For object detection datasets, the classes that do not exist in the image serve as the negative queries. Adding negative detection data can significantly increase 5.4 AP and 7.0 AP$_s$. (3) In this work, we represent each object with a single token to maintain efficiency. We find that increasing the token count per object to 25 brings marginal improvement while inflating the context length by a factor of 25. We hypothesize that the full-image context already preserves sufficient object details, making a single token sufficient to establish an effective correspondence between the object and the global image representation. (4) As LLM adopts causal masks in attention layers, we study whether the proposal order will affect the performance. We observe that positive objects can appear at arbitrary positions within the token sequence during training. As a result, the model develops robustness to proposal ordering, and evaluation performance remains stable even when proposals are randomly shuffled.

\begin{table}[t]
  \centering
  \caption{Ablation studies on WeDetect-Ref. Models are trained with a part of data.}
  \vspace{-0.6em}
  \resizebox{\linewidth}{!}{
      \begin{tabular}{c|l|c|cccc}
        \hline
        \multirow{2}{*}{Exp}  & \multirow{2}{*}{Model}  & RefCOCO & \multicolumn{4}{c}{COCO}  \\
        & & Avg. & AP & AP$_s$ & AP$_m$ & AP$_l$ \\
        \hline
        \rowcolor{ours} & WeDetect-Ref 4B & 93.0 & 47.5 & 29.6 & 54.7 & 67.9 \\
        (1) & BCE loss & 92.8 & 43.1 & 29.0 & 52.5 & 60.7  \\
        (2) & w/o negative det data & 93.1 & 42.1 & 22.6 & 48.0 & 63.7 \\
        (3) & 25 tokens per object & 93.3 & 47.6 & 30.8 & 55.3 & 67.5 \\
        (4) & shuffle proposals & 93.1 & 47.6 & 30.2 & 54.9 & 67.7 \\
        \hline
      \end{tabular}
    }
  \label{tab:ab4}
  \vspace{-0.6em}
\end{table}

\section{Conclusion}

In this work, we propose a family of open-vocabulary detection models following the retrieval methodology, in which targets are simply picked up from a candidate list, rather than generating a query-specific temporary candidate list. This design principle does not use computationally intensive cross-modal fusion layers and ensures a high inference speed. Following the methodology, we propose (1) WeDetect, which is an open-vocabulary object detection foundation, (2) WeDetect-Uni, which is a universal proposal generator and can be applied to a new object retrieval application, and (3) WeDetect-Ref, an LLM-based REC model discarding next-token prediction. The WeDetect model family attains state-of-the-art performance across 15 diverse benchmarks, demonstrating strong generalization ability, and still runs at an extremely fast speed.

{
    \small
    \bibliographystyle{ieeenat_fullname}
    \bibliography{main}

\begin{thebibliography}{90}
\providecommand{\natexlab}[1]{#1}
\providecommand{\url}[1]{\texttt{#1}}
\expandafter\ifx\csname urlstyle\endcsname\relax
  \providecommand{\doi}[1]{doi: #1}\else
  \providecommand{\doi}{doi: \begingroup \urlstyle{rm}\Url}\fi

\bibitem[Bai et~al.(2025{\natexlab{a}})Bai, Cai, Chen, Chen, Chen, Cheng, Deng, Ding, Gao, Ge, Ge, Guo, Huang, Huang, Huang, Hui, Jiang, Li, Li, Li, Li, Lin, Lin, Liu, Liu, Liu, Liu, Liu, Liu, Lu, Luo, Lv, Men, Meng, Ren, Ren, Song, Sun, Tang, Tu, Wan, Wang, Wang, Wang, Wang, Xie, Xu, Xu, Xu, Yang, Yang, Yang, Yang, Yu, Zhang, Zhang, Zhang, Zheng, Zhong, Zhou, Zhou, Zhou, Zhu, and Zhu]{Qwen3-VL}
Shuai Bai, Yuxuan Cai, Ruizhe Chen, Keqin Chen, Xionghui Chen, Zesen Cheng, Lianghao Deng, Wei Ding, Chang Gao, Chunjiang Ge, Wenbin Ge, Zhifang Guo, Qidong Huang, Jie Huang, Fei Huang, Binyuan Hui, Shutong Jiang, Zhaohai Li, Mingsheng Li, Mei Li, Kaixin Li, Zicheng Lin, Junyang Lin, Xuejing Liu, Jiawei Liu, Chenglong Liu, Yang Liu, Dayiheng Liu, Shixuan Liu, Dunjie Lu, Ruilin Luo, Chenxu Lv, Rui Men, Lingchen Meng, Xuancheng Ren, Xingzhang Ren, Sibo Song, Yuchong Sun, Jun Tang, Jianhong Tu, Jianqiang Wan, Peng Wang, Pengfei Wang, Qiuyue Wang, Yuxuan Wang, Tianbao Xie, Yiheng Xu, Haiyang Xu, Jin Xu, Zhibo Yang, Mingkun Yang, Jianxin Yang, An Yang, Bowen Yu, Fei Zhang, Hang Zhang, Xi Zhang, Bo Zheng, Humen Zhong, Jingren Zhou, Fan Zhou, Jing Zhou, Yuanzhi Zhu, and Ke Zhu.
\newblock Qwen3-vl technical report.
\newblock \emph{arXiv preprint arXiv:2511.21631}, 2025{\natexlab{a}}.

\bibitem[Bai et~al.(2025{\natexlab{b}})Bai, Chen, Liu, Wang, Ge, Song, Dang, Wang, Wang, Tang, et~al.]{bai2025qwen25}
Shuai Bai, Keqin Chen, Xuejing Liu, Jialin Wang, Wenbin Ge, Sibo Song, Kai Dang, Peng Wang, Shijie Wang, Jun Tang, et~al.
\newblock Qwen2. 5-vl technical report.
\newblock \emph{arXiv preprint arXiv:2502.13923}, 2025{\natexlab{b}}.

\bibitem[Carion et~al.(2020)Carion, Massa, Synnaeve, Usunier, Kirillov, and Zagoruyko]{detr}
Nicolas Carion, Francisco Massa, Gabriel Synnaeve, Nicolas Usunier, Alexander Kirillov, and Sergey Zagoruyko.
\newblock End-to-end object detection with transformers.
\newblock In \emph{ECCV}, 2020.

\bibitem[Changpinyo et~al.(2021)Changpinyo, Sharma, Ding, and Soricut]{cc12m}
Soravit Changpinyo, Piyush Sharma, Nan Ding, and Radu Soricut.
\newblock Conceptual 12m: Pushing web-scale image-text pre-training to recognize long-tail visual concepts.
\newblock In \emph{CVPR}, 2021.

\bibitem[Chen et~al.(2025)Chen, Wei, Zhao, Song, Wu, Peng, Chan, and Zhang]{ref-l4}
Jierun Chen, Fangyun Wei, Jinjing Zhao, Sizhe Song, Bohuai Wu, Zhuoxuan Peng, S-H~Gary Chan, and Hongyang Zhang.
\newblock Revisiting referring expression comprehension evaluation in the era of large multimodal models.
\newblock In \emph{CVPR}, 2025.

\bibitem[Chen et~al.(2024)Chen, Wang, Cao, Liu, Gao, Cui, Zhu, Ye, Tian, Liu, et~al.]{chen2024expanding}
Zhe Chen, Weiyun Wang, Yue Cao, Yangzhou Liu, Zhangwei Gao, Erfei Cui, Jinguo Zhu, Shenglong Ye, Hao Tian, Zhaoyang Liu, et~al.
\newblock Expanding performance boundaries of open-source multimodal models with model, data, and test-time scaling.
\newblock \emph{arXiv preprint arXiv:2412.05271}, 2024.

\bibitem[Cheng et~al.(2024)Cheng, Song, Ge, Liu, Wang, and Shan]{yolo_world}
Tianheng Cheng, Lin Song, Yixiao Ge, Wenyu Liu, Xinggang Wang, and Ying Shan.
\newblock Yolo-world: Real-time open-vocabulary object detection.
\newblock In \emph{CVPR}, 2024.

\bibitem[Conneau et~al.(2020)Conneau, Khandelwal, Goyal, Chaudhary, Wenzek, Guzm{\'a}n, Grave, Ott, Zettlemoyer, and Stoyanov]{conneau2020unsupervised}
Alexis Conneau, Kartikay Khandelwal, Naman Goyal, Vishrav Chaudhary, Guillaume Wenzek, Francisco Guzm{\'a}n, Edouard Grave, Myle Ott, Luke Zettlemoyer, and Veselin Stoyanov.
\newblock Unsupervised cross-lingual representation learning at scale.
\newblock In \emph{ACL}, 2020.

\bibitem[Dao(2024)]{dao2023flashattention2}
Tri Dao.
\newblock Flash{A}ttention-2: Faster attention with better parallelism and work partitioning.
\newblock In \emph{ICLR}, 2024.

\bibitem[Dave et~al.(2021)Dave, Doll{\'a}r, Ramanan, Kirillov, and Girshick]{dave2021evaluating}
Achal Dave, Piotr Doll{\'a}r, Deva Ramanan, Alexander Kirillov, and Ross Girshick.
\newblock Evaluating large-vocabulary object detectors: The devil is in the details.
\newblock \emph{arXiv preprint arXiv:2102.01066}, 2021.

\bibitem[Deng et~al.(2009)Deng, Dong, Socher, Li, Li, and Fei-Fei]{deng2009imagenet}
Jia Deng, Wei Dong, Richard Socher, Li-Jia Li, Kai Li, and Li Fei-Fei.
\newblock Imagenet: A large-scale hierarchical image database.
\newblock In \emph{CVPR}, 2009.

\bibitem[Dosovitskiy(2020)]{dosovitskiy2020image}
Alexey Dosovitskiy.
\newblock An image is worth 16x16 words: Transformers for image recognition at scale.
\newblock \emph{arXiv preprint arXiv:2010.11929}, 2020.

\bibitem[Dou et~al.(2022)Dou, Kamath, Gan, Zhang, Wang, Li, Liu, Liu, LeCun, Peng, et~al.]{fiber}
Zi-Yi Dou, Aishwarya Kamath, Zhe Gan, Pengchuan Zhang, Jianfeng Wang, Linjie Li, Zicheng Liu, Ce Liu, Yann LeCun, Nanyun Peng, et~al.
\newblock Coarse-to-fine vision-language pre-training with fusion in the backbone.
\newblock In \emph{NeurIPS}, 2022.

\bibitem[Fu et~al.(2023)Fu, Yan, Gao, Xie, and Zheng]{fu2023asag}
Shenghao Fu, Junkai Yan, Yipeng Gao, Xiaohua Xie, and Wei-Shi Zheng.
\newblock Asag: Building strong one-decoder-layer sparse detectors via adaptive sparse anchor generation.
\newblock In \emph{ICCV}, 2023.

\bibitem[Fu et~al.(2024)Fu, Yan, Yang, Wei, Xie, and Zheng]{fu2024frozen-detr}
Shenghao Fu, Junkai Yan, Qize Yang, Xihan Wei, Xiaohua Xie, and Wei-Shi Zheng.
\newblock Frozen-detr: Enhancing detr with image understanding from frozen foundation models.
\newblock In \emph{NeurIPS}, 2024.

\bibitem[Fu et~al.(2025{\natexlab{a}})Fu, Yan, Yang, Wei, Xie, and Zheng]{fu2025HD-OVD}
Shenghao Fu, Junkai Yan, Qize Yang, Xihan Wei, Xiaohua Xie, and Wei-Shi Zheng.
\newblock A hierarchical semantic distillation framework for open-vocabulary object detection.
\newblock \emph{TMM}, 2025{\natexlab{a}}.

\bibitem[Fu et~al.(2025{\natexlab{b}})Fu, Yang, Mo, Yan, Wei, Meng, Xie, and Zheng]{fu2025llmdet}
Shenghao Fu, Qize Yang, Qijie Mo, Junkai Yan, Xihan Wei, Jingke Meng, Xiaohua Xie, and Wei-Shi Zheng.
\newblock Llmdet: Learning strong open-vocabulary object detectors under the supervision of large language models.
\newblock In \emph{CVPR}, 2025{\natexlab{b}}.

\bibitem[Gu et~al.(2021)Gu, Lin, Kuo, and Cui]{vild}
Xiuye Gu, Tsung-Yi Lin, Weicheng Kuo, and Yin Cui.
\newblock Open-vocabulary object detection via vision and language knowledge distillation.
\newblock In \emph{ICLR}, 2021.

\bibitem[Guan et~al.(2025)Guan, Sun, Fu, Huang, Yuan, and Li]{prompt-dino}
Yuchen Guan, Chong Sun, Canmiao Fu, Zhipeng Huang, Chun Yuan, and Chen Li.
\newblock Text-guided visual prompt dino for generic segmentation.
\newblock In \emph{ICCV}, 2025.

\bibitem[Gupta et~al.(2019)Gupta, Dollar, and Girshick]{gupta2019lvis}
Agrim Gupta, Piotr Dollar, and Ross Girshick.
\newblock Lvis: A dataset for large vocabulary instance segmentation.
\newblock In \emph{CVPR}, 2019.

\bibitem[Hao et~al.(2025)Hao, Zhu, Guo, Guo, Jiang, Lu, Tang, and Wang]{hao2025referring}
Xiangzhao Hao, Kuan Zhu, Hongyu Guo, Haiyun Guo, Ning Jiang, Quan Lu, Ming Tang, and Jinqiao Wang.
\newblock Referring expression instance retrieval and a strong end-to-end baseline.
\newblock In \emph{ACM MM}, 2025.

\bibitem[He et~al.(2016)He, Zhang, Ren, and Sun]{resnet}
Kaiming He, Xiangyu Zhang, Shaoqing Ren, and Jian Sun.
\newblock Deep residual learning for image recognition.
\newblock In \emph{CVPR}, 2016.

\bibitem[Jiang et~al.(2024{\natexlab{a}})Jiang, Li, Zeng, Ren, Liu, and Zhang]{T-Rex2}
Qing Jiang, Feng Li, Zhaoyang Zeng, Tianhe Ren, Shilong Liu, and Lei Zhang.
\newblock T-rex2: Towards generic object detection via text-visual prompt synergy.
\newblock In \emph{ECCV}, 2024{\natexlab{a}}.

\bibitem[Jiang et~al.(2024{\natexlab{b}})Jiang, Luo, Yang, Xiong, Chen, Zeng, Ren, and Zhang]{jiang2024chatrex}
Qing Jiang, Gen Luo, Yuqin Yang, Yuda Xiong, Yihao Chen, Zhaoyang Zeng, Tianhe Ren, and Lei Zhang.
\newblock Chatrex: Taming multimodal llm for joint perception and understanding.
\newblock \emph{arXiv preprint arXiv:2411.18363}, 2024{\natexlab{b}}.

\bibitem[Jiang et~al.(2025{\natexlab{a}})Jiang, Huo, Chen, Xiong, Zeng, Chen, Ren, Yu, and Zhang]{jiang2025detect}
Qing Jiang, Junan Huo, Xingyu Chen, Yuda Xiong, Zhaoyang Zeng, Yihao Chen, Tianhe Ren, Junzhi Yu, and Lei Zhang.
\newblock Detect anything via next point prediction.
\newblock \emph{arXiv preprint arXiv:2510.12798}, 2025{\natexlab{a}}.

\bibitem[Jiang et~al.(2025{\natexlab{b}})Jiang, Wu, Zeng, Ren, Xiong, Chen, Qin, and Zhang]{jiang2025referring}
Qing Jiang, Lin Wu, Zhaoyang Zeng, Tianhe Ren, Yuda Xiong, Yihao Chen, Liu Qin, and Lei Zhang.
\newblock Referring to any person.
\newblock In \emph{ICCV}, 2025{\natexlab{b}}.

\bibitem[Jiao et~al.(2023)Jiao, Tang, Lin, Gao, Ma, Wang, and Zheng]{jiao2023dilateformer}
Jiayu Jiao, Yu-Ming Tang, Kun-Yu Lin, Yipeng Gao, Andy~J Ma, Yaowei Wang, and Wei-Shi Zheng.
\newblock Dilateformer: Multi-scale dilated transformer for visual recognition.
\newblock \emph{TMM}, 2023.

\bibitem[Kamath et~al.(2021)Kamath, Singh, LeCun, Synnaeve, Misra, and Carion]{kamath2021mdetr}
Aishwarya Kamath, Mannat Singh, Yann LeCun, Gabriel Synnaeve, Ishan Misra, and Nicolas Carion.
\newblock Mdetr-modulated detection for end-to-end multi-modal understanding.
\newblock In \emph{ICCV}, 2021.

\bibitem[Kazemzadeh et~al.(2014)Kazemzadeh, Ordonez, Matten, and Berg]{refcoco}
Sahar Kazemzadeh, Vicente Ordonez, Mark Matten, and Tamara Berg.
\newblock Referitgame: Referring to objects in photographs of natural scenes.
\newblock In \emph{EMNLP}, 2014.

\bibitem[Kim et~al.(2022)Kim, Lin, Angelova, Kweon, and Kuo]{oln}
Dahun Kim, Tsung-Yi Lin, Anelia Angelova, In~So Kweon, and Weicheng Kuo.
\newblock Learning open-world object proposals without learning to classify.
\newblock \emph{IEEE Robotics and Automation Letters}, 2022.

\bibitem[Kirillov et~al.(2023)Kirillov, Mintun, Ravi, Mao, Rolland, Gustafson, Xiao, Whitehead, Berg, Lo, et~al.]{sam}
Alexander Kirillov, Eric Mintun, Nikhila Ravi, Hanzi Mao, Chloe Rolland, Laura Gustafson, Tete Xiao, Spencer Whitehead, Alexander~C Berg, Wan-Yen Lo, et~al.
\newblock Segment anything.
\newblock In \emph{ICCV}, 2023.

\bibitem[Krishna et~al.(2017)Krishna, Zhu, Groth, Johnson, Hata, Kravitz, Chen, Kalantidis, Li, Shamma, et~al.]{vg}
Ranjay Krishna, Yuke Zhu, Oliver Groth, Justin Johnson, Kenji Hata, Joshua Kravitz, Stephanie Chen, Yannis Kalantidis, Li-Jia Li, David~A Shamma, et~al.
\newblock Visual genome: Connecting language and vision using crowdsourced dense image annotations.
\newblock \emph{IJCV}, 2017.

\bibitem[Kuznetsova et~al.(2020)Kuznetsova, Rom, Alldrin, Uijlings, Krasin, Pont-Tuset, Kamali, Popov, Malloci, Kolesnikov, et~al.]{kuznetsova2020open}
Alina Kuznetsova, Hassan Rom, Neil Alldrin, Jasper Uijlings, Ivan Krasin, Jordi Pont-Tuset, Shahab Kamali, Stefan Popov, Matteo Malloci, Alexander Kolesnikov, et~al.
\newblock The open images dataset v4: Unified image classification, object detection, and visual relationship detection at scale.
\newblock \emph{IJCV}, 2020.

\bibitem[Lai et~al.(2024)Lai, Tian, Chen, Li, Yuan, Liu, and Jia]{lai2024lisa}
Xin Lai, Zhuotao Tian, Yukang Chen, Yanwei Li, Yuhui Yuan, Shu Liu, and Jiaya Jia.
\newblock Lisa: Reasoning segmentation via large language model.
\newblock In \emph{CVPR}, 2024.

\bibitem[Li et~al.(2024)Li, Zhang, Guo, Zhang, Li, Zhang, Zhang, Li, Liu, and Li]{llava-onevision}
Bo Li, Yuanhan Zhang, Dong Guo, Renrui Zhang, Feng Li, Hao Zhang, Kaichen Zhang, Yanwei Li, Ziwei Liu, and Chunyuan Li.
\newblock Llava-onevision: Easy visual task transfer.
\newblock \emph{arXiv preprint arXiv:2408.03326}, 2024.

\bibitem[Li et~al.(2022{\natexlab{a}})Li, Liu, Li, Zhang, Aneja, Yang, Jin, Hu, Liu, Lee, et~al.]{li2022elevater}
Chunyuan Li, Haotian Liu, Liunian Li, Pengchuan Zhang, Jyoti Aneja, Jianwei Yang, Ping Jin, Houdong Hu, Zicheng Liu, Yong~Jae Lee, et~al.
\newblock Elevater: A benchmark and toolkit for evaluating language-augmented visual models.
\newblock In \emph{NeurIPS}, 2022{\natexlab{a}}.

\bibitem[Li et~al.(2023{\natexlab{a}})Li, Li, Geng, Jiang, Cheng, Zhang, Ke, Xu, and Chu]{li2023yolov6v3}
Chuyi Li, Lulu Li, Yifei Geng, Hongliang Jiang, Meng Cheng, Bo Zhang, Zaidan Ke, Xiaoming Xu, and Xiangxiang Chu.
\newblock Yolov6 v3. 0: A full-scale reloading.
\newblock \emph{arXiv preprint arXiv:2301.05586}, 2023{\natexlab{a}}.

\bibitem[Li et~al.(2025)Li, Xie, Ao, Leng, and Yin]{li2025lmm}
Jincheng Li, Chunyu Xie, Ji Ao, Dawei Leng, and Yuhui Yin.
\newblock Lmm-det: Make large multimodal models excel in object detection.
\newblock In \emph{ICCV}, 2025.

\bibitem[Li et~al.(2023{\natexlab{b}})Li, Dou, Peng, and Chang]{li2024desco}
Liunian Li, Zi-Yi Dou, Nanyun Peng, and Kai-Wei Chang.
\newblock Desco: Learning object recognition with rich language descriptions.
\newblock In \emph{NeurIPS}, 2023{\natexlab{b}}.

\bibitem[Li et~al.(2022{\natexlab{b}})Li, Zhang, Zhang, Yang, Li, Zhong, Wang, Yuan, Zhang, Hwang, et~al.]{GLIP}
Liunian~Harold Li, Pengchuan Zhang, Haotian Zhang, Jianwei Yang, Chunyuan Li, Yiwu Zhong, Lijuan Wang, Lu Yuan, Lei Zhang, Jenq-Neng Hwang, et~al.
\newblock Grounded language-image pre-training.
\newblock In \emph{CVPR}, 2022{\natexlab{b}}.

\bibitem[Lian et~al.(2025)Lian, Ding, Ge, Liu, Mao, Li, Pavone, Liu, Darrell, Yala, et~al.]{dam}
Long Lian, Yifan Ding, Yunhao Ge, Sifei Liu, Hanzi Mao, Boyi Li, Marco Pavone, Ming-Yu Liu, Trevor Darrell, Adam Yala, et~al.
\newblock Describe anything: Detailed localized image and video captioning.
\newblock In \emph{ICCV}, 2025.

\bibitem[Lin et~al.(2014)Lin, Maire, Belongie, Hays, Perona, Ramanan, Doll{\'a}r, and Zitnick]{coco}
Tsung-Yi Lin, Michael Maire, Serge Belongie, James Hays, Pietro Perona, Deva Ramanan, Piotr Doll{\'a}r, and C~Lawrence Zitnick.
\newblock Microsoft coco: Common objects in context.
\newblock In \emph{ECCV}, 2014.

\bibitem[Lin et~al.(2017)Lin, Goyal, Girshick, He, and Doll{\'a}r]{ross2017focal}
Tsung-Yi Lin, Priya Goyal, Ross Girshick, Kaiming He, and Piotr Doll{\'a}r.
\newblock Focal loss for dense object detection.
\newblock In \emph{CVPR}, 2017.

\bibitem[Liu et~al.(2023{\natexlab{a}})Liu, Ding, and Jiang]{grefcoco}
Chang Liu, Henghui Ding, and Xudong Jiang.
\newblock Gres: Generalized referring expression segmentation.
\newblock In \emph{CVPR}, 2023{\natexlab{a}}.

\bibitem[Liu et~al.(2023{\natexlab{b}})Liu, Li, Wu, and Lee]{llava}
Haotian Liu, Chunyuan Li, Qingyang Wu, and Yong~Jae Lee.
\newblock Visual instruction tuning.
\newblock In \emph{NeurIPS}, 2023{\natexlab{b}}.

\bibitem[Liu et~al.(2024{\natexlab{a}})Liu, Yang, Li, and Wang]{liu2024finecops}
Junzhuo Liu, Xuzheng Yang, Weiwei Li, and Peng Wang.
\newblock Finecops-ref: A new dataset and task for fine-grained compositional referring expression comprehension.
\newblock \emph{arXiv preprint arXiv:2409.14750}, 2024{\natexlab{a}}.

\bibitem[Liu et~al.(2025)Liu, Shen, Fang, Sun, Liao, and Zhao]{liu2025vlm}
Peng Liu, Haozhan Shen, Chunxin Fang, Zhicheng Sun, Jiajia Liao, and Tiancheng Zhao.
\newblock Vlm-fo1: Bridging the gap between high-level reasoning and fine-grained perception in vlms.
\newblock \emph{arXiv preprint arXiv:2509.25916}, 2025.

\bibitem[Liu et~al.(2024{\natexlab{b}})Liu, Zeng, Ren, Li, Zhang, Yang, Li, Yang, Su, Zhu, et~al.]{grounding_dino}
Shilong Liu, Zhaoyang Zeng, Tianhe Ren, Feng Li, Hao Zhang, Jie Yang, Chunyuan Li, Jianwei Yang, Hang Su, Jun Zhu, et~al.
\newblock Grounding dino: Marrying dino with grounded pre-training for open-set object detection.
\newblock In \emph{ECCV}, 2024{\natexlab{b}}.

\bibitem[Liu et~al.(2022)Liu, Mao, Wu, Feichtenhofer, Darrell, and Xie]{liu2022convnext}
Zhuang Liu, Hanzi Mao, Chao-Yuan Wu, Christoph Feichtenhofer, Trevor Darrell, and Saining Xie.
\newblock A convnet for the 2020s.
\newblock In \emph{CVPR}, 2022.

\bibitem[Long et~al.(2023)Long, Wen, Han, Xu, Ren, Zhang, Zhao, and Liang]{long2023capdet}
Yanxin Long, Youpeng Wen, Jianhua Han, Hang Xu, Pengzhen Ren, Wei Zhang, Shen Zhao, and Xiaodan Liang.
\newblock Capdet: Unifying dense captioning and open-world detection pretraining.
\newblock In \emph{CVPR}, 2023.

\bibitem[Ma et~al.(2024)Ma, Jiang, Wu, Yuan, and Qi]{ma2024groma}
Chuofan Ma, Yi Jiang, Jiannan Wu, Zehuan Yuan, and Xiaojuan Qi.
\newblock Groma: Localized visual tokenization for grounding multimodal large language models.
\newblock In \emph{ECCV}, 2024.

\bibitem[Maaz et~al.(2022)Maaz, Rasheed, Khan, Khan, Anwer, and Yang]{mavl}
Muhammad Maaz, Hanoona Rasheed, Salman Khan, Fahad~Shahbaz Khan, Rao~Muhammad Anwer, and Ming-Hsuan Yang.
\newblock Class-agnostic object detection with multi-modal transformer.
\newblock In \emph{ECCV}, 2022.

\bibitem[Mao et~al.(2016)Mao, Huang, Toshev, Camburu, Yuille, and Murphy]{refcocog}
Junhua Mao, Jonathan Huang, Alexander Toshev, Oana Camburu, Alan~L Yuille, and Kevin Murphy.
\newblock Generation and comprehension of unambiguous object descriptions.
\newblock In \emph{CVPR}, 2016.

\bibitem[Mao et~al.(2023)Mao, Chen, Zhu, Chen, Su, Zhang, and Xue]{mao2023coco}
Xiaofeng Mao, Yuefeng Chen, Yao Zhu, Da Chen, Hang Su, Rong Zhang, and Hui Xue.
\newblock Coco-o: A benchmark for object detectors under natural distribution shifts.
\newblock In \emph{ICCV}, 2023.

\bibitem[Minderer et~al.(2023)Minderer, Gritsenko, and Houlsby]{OWL-ST}
Matthias Minderer, Alexey Gritsenko, and Neil Houlsby.
\newblock Scaling open-vocabulary object detection.
\newblock In \emph{NeurIPS}, 2023.

\bibitem[Park et~al.(2024)Park, Saito, and Kim]{park2024weak}
Kwanyong Park, Kuniaki Saito, and Donghyun Kim.
\newblock Weak-to-strong compositional learning from generative models for language-based object detection.
\newblock In \emph{ECCV}, 2024.

\bibitem[Plummer et~al.(2015)Plummer, Wang, Cervantes, Caicedo, Hockenmaier, and Lazebnik]{plummer2015flickr30k}
Bryan~A Plummer, Liwei Wang, Chris~M Cervantes, Juan~C Caicedo, Julia Hockenmaier, and Svetlana Lazebnik.
\newblock Flickr30k entities: Collecting region-to-phrase correspondences for richer image-to-sentence models.
\newblock In \emph{CVPR}, 2015.

\bibitem[Radford et~al.(2021)Radford, Kim, Hallacy, Ramesh, Goh, Agarwal, Sastry, Askell, Mishkin, Clark, et~al.]{clip}
Alec Radford, Jong~Wook Kim, Chris Hallacy, Aditya Ramesh, Gabriel Goh, Sandhini Agarwal, Girish Sastry, Amanda Askell, Pamela Mishkin, Jack Clark, et~al.
\newblock Learning transferable visual models from natural language supervision.
\newblock In \emph{ICML}, 2021.

\bibitem[Ramanathan et~al.(2023)Ramanathan, Kalia, Petrovic, Wen, Zheng, Guo, Wang, Marquez, Kovvuri, Kadian, et~al.]{ramanathan2023paco}
Vignesh Ramanathan, Anmol Kalia, Vladan Petrovic, Yi Wen, Baixue Zheng, Baishan Guo, Rui Wang, Aaron Marquez, Rama Kovvuri, Abhishek Kadian, et~al.
\newblock Paco: Parts and attributes of common objects.
\newblock In \emph{CVPR}, 2023.

\bibitem[Ren et~al.(2016)Ren, He, Girshick, and Sun]{ren2016faster}
Shaoqing Ren, Kaiming He, Ross Girshick, and Jian Sun.
\newblock Faster r-cnn: Towards real-time object detection with region proposal networks.
\newblock \emph{TPAMI}, 2016.

\bibitem[Ren et~al.(2024)Ren, Chen, Jiang, Zeng, Xiong, Liu, Ma, Shen, Gao, Jiang, et~al.]{ren2024dinox}
Tianhe Ren, Yihao Chen, Qing Jiang, Zhaoyang Zeng, Yuda Xiong, Wenlong Liu, Zhengyu Ma, Junyi Shen, Yuan Gao, Xiaoke Jiang, et~al.
\newblock Dino-x: A unified vision model for open-world object detection and understanding.
\newblock \emph{arXiv preprint arXiv:2411.14347}, 2024.

\bibitem[Schuhmann et~al.(2021)Schuhmann, Vencu, Beaumont, Kaczmarczyk, Mullis, Katta, Coombes, Jitsev, and Komatsuzaki]{schuhmann2021laion}
Christoph Schuhmann, Richard Vencu, Romain Beaumont, Robert Kaczmarczyk, Clayton Mullis, Aarush Katta, Theo Coombes, Jenia Jitsev, and Aran Komatsuzaki.
\newblock Laion-400m: Open dataset of clip-filtered 400 million image-text pairs.
\newblock \emph{arXiv preprint arXiv:2111.02114}, 2021.

\bibitem[Shao et~al.(2019)Shao, Li, Zhang, Peng, Yu, Zhang, Li, and Sun]{shao2019objects365}
Shuai Shao, Zeming Li, Tianyuan Zhang, Chao Peng, Gang Yu, Xiangyu Zhang, Jing Li, and Jian Sun.
\newblock Objects365: A large-scale, high-quality dataset for object detection.
\newblock In \emph{ICCV}, 2019.

\bibitem[Shen et~al.(2025)Shen, Liu, Li, Fang, Ma, Liao, Shen, Zhang, Zhao, Zhang, et~al.]{shen2025vlmr1}
Haozhan Shen, Peng Liu, Jingcheng Li, Chunxin Fang, Yibo Ma, Jiajia Liao, Qiaoli Shen, Zilun Zhang, Kangjia Zhao, Qianqian Zhang, et~al.
\newblock Vlm-r1: A stable and generalizable r1-style large vision-language model.
\newblock \emph{arXiv preprint arXiv:2504.07615}, 2025.

\bibitem[Su et~al.(2025)Su, Zhang, Li, Liu, Liao, Pan, Liu, Xing, Sun, Li, et~al.]{su2025patch}
Yongyi Su, Haojie Zhang, Shijie Li, Nanqing Liu, Jingyi Liao, Junyi Pan, Yuan Liu, Xiaofen Xing, Chong Sun, Chen Li, et~al.
\newblock Patch-as-decodable-token: Towards unified multi-modal vision tasks in mllms.
\newblock \emph{arXiv preprint arXiv:2510.01954}, 2025.

\bibitem[Tang et~al.(2025)Tang, Xie, Wang, Bao, Weng, Li, Zheng, and Wang]{tang2025ufo}
Hao Tang, Chenwei Xie, Haiyang Wang, Xiaoyi Bao, Tingyu Weng, Pandeng Li, Yun Zheng, and Liwei Wang.
\newblock Ufo: A unified approach to fine-grained visual perception via open-ended language interface.
\newblock In \emph{NeurIPS}, 2025.

\bibitem[Wang et~al.(2025{\natexlab{a}})Wang, Liu, Chen, Lin, Han, and Ding]{wang2025yoloe}
Ao Wang, Lihao Liu, Hui Chen, Zijia Lin, Jungong Han, and Guiguang Ding.
\newblock Yoloe: Real-time seeing anything.
\newblock In \emph{ICCV}, 2025{\natexlab{a}}.

\bibitem[Wang et~al.(2024)Wang, Ren, Jie, Dong, Feng, Qian, Ma, Jiang, Wang, Lan, et~al.]{ov-dino}
Hao Wang, Pengzhen Ren, Zequn Jie, Xiao Dong, Chengjian Feng, Yinlong Qian, Lin Ma, Dongmei Jiang, Yaowei Wang, Xiangyuan Lan, et~al.
\newblock Ov-dino: Unified open-vocabulary detection with language-aware selective fusion.
\newblock \emph{arXiv preprint arXiv:2407.07844}, 2024.

\bibitem[Wang et~al.(2023{\natexlab{a}})Wang, Zhang, Chu, Cao, Zhou, Wu, Wang, He, and Lin]{wang2023v3det}
Jiaqi Wang, Pan Zhang, Tao Chu, Yuhang Cao, Yujie Zhou, Tong Wu, Bin Wang, Conghui He, and Dahua Lin.
\newblock V3det: Vast vocabulary visual detection dataset.
\newblock In \emph{ICCV}, 2023{\natexlab{a}}.

\bibitem[Wang et~al.(2023{\natexlab{b}})Wang, Liu, Du, Ding, Liao, Qi, Chen, and Liu]{oadp}
Luting Wang, Yi Liu, Penghui Du, Zihan Ding, Yue Liao, Qiaosong Qi, Biaolong Chen, and Si Liu.
\newblock Object-aware distillation pyramid for open-vocabulary object detection.
\newblock In \emph{CVPR}, 2023{\natexlab{b}}.

\bibitem[Wang et~al.(2025{\natexlab{b}})Wang, Chen, Huang, Li, Li, Li, Alvarez, Zhang, and Yu]{wang2025videoitg}
Shihao Wang, Guo Chen, De-an Huang, Zhiqi Li, Minghan Li, Guilin Li, Jose~M Alvarez, Lei Zhang, and Zhiding Yu.
\newblock Videoitg: Multimodal video understanding with instructed temporal grounding.
\newblock \emph{arXiv preprint arXiv:2507.13353}, 2025{\natexlab{b}}.

\bibitem[Wang et~al.(2025{\natexlab{c}})Wang, Gao, Gu, Pu, Cui, Wei, Liu, Jing, Ye, Shao, et~al.]{wang2025internvl35}
Weiyun Wang, Zhangwei Gao, Lixin Gu, Hengjun Pu, Long Cui, Xingguang Wei, Zhaoyang Liu, Linglin Jing, Shenglong Ye, Jie Shao, et~al.
\newblock Internvl3. 5: Advancing open-source multimodal models in versatility, reasoning, and efficiency.
\newblock \emph{arXiv preprint arXiv:2508.18265}, 2025{\natexlab{c}}.

\bibitem[Wei et~al.(2025)Wei, Wang, Ma, Mei, Chen, Jin, and Rao]{wei2025hqclip}
Zhixiang Wei, Guangting Wang, Xiaoxiao Ma, Ke Mei, Huaian Chen, Yi Jin, and Fengyun Rao.
\newblock Hq-clip: Leveraging large vision-language models to create high-quality image-text datasets and clip models.
\newblock In \emph{ICCV}, 2025.

\bibitem[Xie et~al.(2023{\natexlab{a}})Xie, Cai, Li, Kong, Wu, Song, Morimitsu, Yao, Wang, Zhang, et~al.]{xie2023ccmb}
Chunyu Xie, Heng Cai, Jincheng Li, Fanjing Kong, Xiaoyu Wu, Jianfei Song, Henrique Morimitsu, Lin Yao, Dexin Wang, Xiangzheng Zhang, et~al.
\newblock Ccmb: A large-scale chinese cross-modal benchmark.
\newblock In \emph{ACM MM}, 2023{\natexlab{a}}.

\bibitem[Xie et~al.(2023{\natexlab{b}})Xie, Zhang, Wu, Zhu, Zhao, and Liang]{d3}
Chi Xie, Zhao Zhang, Yixuan Wu, Feng Zhu, Rui Zhao, and Shuang Liang.
\newblock Described object detection: Liberating object detection with flexible expressions.
\newblock In \emph{NeurIPS}, 2023{\natexlab{b}}.

\bibitem[Xie et~al.(2025)Xie, Wang, Kong, Li, Liang, Ao, Leng, and Yin]{xie2025fgclip2}
Chunyu Xie, Bin Wang, Fanjing Kong, Jincheng Li, Dawei Liang, Ji Ao, Dawei Leng, and Yuhui Yin.
\newblock Fg-clip 2: A bilingual fine-grained vision-language alignment model.
\newblock \emph{arXiv preprint arXiv:2510.10921}, 2025.

\bibitem[Yao et~al.(2022)Yao, Han, Wen, Liang, Xu, Zhang, Li, Xu, and Xu]{yao2022detclip}
Lewei Yao, Jianhua Han, Youpeng Wen, Xiaodan Liang, Dan Xu, Wei Zhang, Zhenguo Li, Chunjing Xu, and Hang Xu.
\newblock Detclip: Dictionary-enriched visual-concept paralleled pre-training for open-world detection.
\newblock In \emph{NeurIPS}, 2022.

\bibitem[Yao et~al.(2023)Yao, Han, Liang, Xu, Zhang, Li, and Xu]{yao2023detclipv2}
Lewei Yao, Jianhua Han, Xiaodan Liang, Dan Xu, Wei Zhang, Zhenguo Li, and Hang Xu.
\newblock Detclipv2: Scalable open-vocabulary object detection pre-training via word-region alignment.
\newblock In \emph{CVPR}, 2023.

\bibitem[Yao et~al.(2024)Yao, Pi, Han, Liang, Xu, Zhang, Li, and Xu]{yao2024detclipv3}
Lewei Yao, Renjie Pi, Jianhua Han, Xiaodan Liang, Hang Xu, Wei Zhang, Zhenguo Li, and Dan Xu.
\newblock Detclipv3: Towards versatile generative open-vocabulary object detection.
\newblock In \emph{CVPR}, 2024.

\bibitem[Yin et~al.(2025)Yin, Ren, Yan, Ding, and Hao]{yin2025rod}
Heng Yin, Yuqiang Ren, Ke Yan, Shouhong Ding, and Yongtao Hao.
\newblock Rod-mllm: Towards more reliable object detection in multimodal large language models.
\newblock In \emph{CVPR}, 2025.

\bibitem[Yu et~al.(2016)Yu, Poirson, Yang, Berg, and Berg]{refcoco+}
Licheng Yu, Patrick Poirson, Shan Yang, Alexander~C Berg, and Tamara~L Berg.
\newblock Modeling context in referring expressions.
\newblock In \emph{ECCV}, 2016.

\bibitem[Yuan et~al.(2024)Yuan, Li, Liu, Tang, Luo, Qin, Zhang, and Zhu]{yuan2024osprey}
Yuqian Yuan, Wentong Li, Jian Liu, Dongqi Tang, Xinjie Luo, Chi Qin, Lei Zhang, and Jianke Zhu.
\newblock Osprey: Pixel understanding with visual instruction tuning.
\newblock In \emph{CVPR}, 2024.

\bibitem[Yuan et~al.(2025)Yuan, Zhang, Li, Cheng, Zhang, Li, Li, Zhao, Zhang, Zhuang, et~al.]{yuan2025videorefer}
Yuqian Yuan, Hang Zhang, Wentong Li, Zesen Cheng, Boqiang Zhang, Long Li, Xin Li, Deli Zhao, Wenqiao Zhang, Yueting Zhuang, et~al.
\newblock Videorefer suite: Advancing spatial-temporal object understanding with video llm.
\newblock In \emph{CVPR}, 2025.

\bibitem[Zellers et~al.(2019)Zellers, Bisk, Farhadi, and Choi]{vcr}
Rowan Zellers, Yonatan Bisk, Ali Farhadi, and Yejin Choi.
\newblock From recognition to cognition: Visual commonsense reasoning.
\newblock In \emph{CVPR}, 2019.

\bibitem[Zhan et~al.(2024)Zhan, Zhu, Chen, Yang, Tang, and Wang]{zhan2024griffon}
Yufei Zhan, Yousong Zhu, Zhiyang Chen, Fan Yang, Ming Tang, and Jinqiao Wang.
\newblock Griffon: Spelling out all object locations at any granularity with large language models.
\newblock In \emph{ECCV}, 2024.

\bibitem[Zhang et~al.(2022)Zhang, Zhang, Hu, Chen, Li, Dai, Wang, Yuan, Hwang, and Gao]{zhang2022glipv2}
Haotian Zhang, Pengchuan Zhang, Xiaowei Hu, Yen-Chun Chen, Liunian Li, Xiyang Dai, Lijuan Wang, Lu Yuan, Jenq-Neng Hwang, and Jianfeng Gao.
\newblock Glipv2: Unifying localization and vision-language understanding.
\newblock In \emph{NeurIPS}, 2022.

\bibitem[Zhang et~al.(2024)Zhang, Sun, Chen, Xiao, Shao, Zhang, Liu, Chen, and Luo]{zhang2024gpt4roi}
Shilong Zhang, Peize Sun, Shoufa Chen, Min Xiao, Wenqi Shao, Wenwei Zhang, Yu Liu, Kai Chen, and Ping Luo.
\newblock Gpt4roi: Instruction tuning large language model on region-of-interest.
\newblock In \emph{ECCV}, 2024.

\bibitem[Zhao et~al.(2024{\natexlab{a}})Zhao, Song, Chen, Rong, Feng, Zhang, Ji, Wang, Ding, and Sun]{zhao2024octopus}
Chuyang Zhao, YuXin Song, Junru Chen, Kang Rong, Haocheng Feng, Gang Zhang, Shufan Ji, Jingdong Wang, Errui Ding, and Yifan Sun.
\newblock Octopus: A multi-modal llm with parallel recognition and sequential understanding.
\newblock In \emph{NeurIPS}, 2024{\natexlab{a}}.

\bibitem[Zhao et~al.(2024{\natexlab{b}})Zhao, Zhao, Suh, Metaxas, Chandraker, Schulter, et~al.]{zhao2024generating}
Shiyu Zhao, Long Zhao, Yumin Suh, Dimitris~N Metaxas, Manmohan Chandraker, Samuel Schulter, et~al.
\newblock Generating enhanced negatives for training language-based object detectors.
\newblock In \emph{CVPR}, 2024{\natexlab{b}}.

\bibitem[Zhao et~al.(2024{\natexlab{c}})Zhao, Chen, Xu, Li, Wang, Li, and Huang]{mm_GDINO}
Xiangyu Zhao, Yicheng Chen, Shilin Xu, Xiangtai Li, Xinjiang Wang, Yining Li, and Haian Huang.
\newblock An open and comprehensive pipeline for unified object grounding and detection.
\newblock \emph{arXiv preprint arXiv:2401.02361}, 2024{\natexlab{c}}.

\end{thebibliography}
}


\maketitlesupplementary

\section{Limitation}

Different from other LMMs with next-token prediction, WeDetect-Ref is a binary classification model, which does not support detecting multiple queries in a single forward pass. Instead, we can detect multiple queries in multiple times and then merge the results, similar to the inference pipeline of Grounding-DINO on LVIS. High performance on COCO and OdinW13 in Table 5 demonstrates the effectiveness of this mechanism. Further, as WeDetect-Ref enjoys a 13x speedup compared to Qwen3-VL, WeDetect-Ref can still run faster than it with a few queries while getting higher performance.

\section{Implementation Details}

\begin{table}[t]
  \centering
  \caption{WeDetect training dataset.}
  \resizebox{\linewidth}{!}{
      \begin{tabular}{c|c}
        \hline
        \#Sample & Dataset \\
        \hline
        \multirow{3}{*}{5.0M} & OpenImagesV6 (1.74M)~\cite{kuznetsova2020open}, Objects365 V2  (1.74M)~\cite{shao2019objects365}\\
        &V3Det (0.18M)~\cite{wang2023v3det}, ImageNetBox (0.54M)~\cite{deng2009imagenet} \\
        &  GoldG (0.8M)~\cite{GLIP}  \\
        \hline
        15.0M & self-collected data (15.0M) \\
        \hline
      \end{tabular}
    }
  \label{tab:data1}
\end{table}

\begin{table}[t]
  \centering
  \caption{WeDetect-Ref training dataset.}
  \resizebox{\linewidth}{!}{
      \begin{tabular}{l|c|c|c}
        \hline
        Stage & \#Sample & Task & Dataset \\
        \hline
        \multirow{2}{*}{Stage1} & \multirow{2}{*}{685k} & image caption  & LLaVA Pretrain (558k)~\cite{llava} \\
        \cline{3-4}
        & & region caption & refcoco/+/g (127k)~\cite{refcoco, refcoco+, refcocog} \\
        \hline
        \multirow{4}{*}{Stage2} & \multirow{4}{*}{1685k} & image QA & LLaVA-OneVision-SI (835k)~\cite{llava-onevision} \\
        \cline{3-4}
        & & grounded conversation & VCR (213k)~\cite{vcr}, Osprey (145k)~\cite{yuan2024osprey} \\
        \cline{3-4}
        & & \multirow{2}{*}{region caption} & refcoco/+/g (56k)~\cite{refcoco, refcoco+, refcocog} \\
        & & & DAM-LVIS (90k)~\cite{dam}, VG (346k)~\cite{vg} \\
        \hline
        \multirow{6}{*}{Stage3} & \multirow{6}{*}{3911k} & \multirow{2}{*}{detection} & V3Det (1122k)~\cite{wang2023v3det}, LVIS (661k)~\cite{gupta2019lvis} \\
        & & & COCO (698k)~\cite{coco} \\
        \cline{3-4}
        & & \multirow{4}{*}{REC} & refcoco/+/g (127k)~\cite{refcoco, refcoco+, refcocog}, flickr (185k)~\cite{plummer2015flickr30k} \\
        & & & humanref (135k)~\cite{jiang2025referring}, Ref-L4 (45k)~\cite{ref-l4} \\
        & & & grefcoco (209k)~\cite{grefcoco}, FineCops-Ref (182k)~\cite{liu2024finecops} \\
        & & & reircoco (547k)~\cite{hao2025referring} \\
        \hline
      \end{tabular}
    }
  \label{tab:data2}
\end{table}

\begin{table}[t]
  \centering
  \caption{WeDetect-Ref training settings.}
  \resizebox{\linewidth}{!}{
      \begin{tabular}{l|c|c|c}
        \hline
        Config & Stage1 & Stage2 & Stage3 \\
        \hline
        training module & projector & projector + LLM & projector + LLM \\
        training objective & language modeling loss & language modeling loss & sigmoid focal loss \\
        batch size & 256 & 128 & 128 \\
        learning rate & $1e^{-3}$ & $1e^{-5}$ & $1e^{-5}$ \\
        epoch & 1 & 1 & 1\\
        \hline
      \end{tabular}
    }
  \label{tab:config}
\end{table}

WeDetect is trained with around 20M samples, including 15M self-collected data and 5M open-sourced data. Details are summarized in \Cref{tab:data1}. Each class name is encoded separately and is represented as a single embedding. For head and neck initialization, the model is trained for 20 epochs with a learning rate of $5e^{-4}$. For end-to-end training, the model is trained for 30 epochs with a learning rate of $1e^{-5}$. The total batch size is 320.

WeDetect-Ref is finetuned from Qwen3-VL~\cite{Qwen3-VL} using a three-stage training strategy. The training data and configurations are summarized in \Cref{tab:data2} and \Cref{tab:config}, respectively. As shown in Table 7, we find that incorporating negative detection data is crucial for achieving high detection performance. To this end, for each image in the detection datasets, we select three negative class names as negative training samples. These negative class names are chosen based on the top-3 confidence predictions from WeDetect among the negative classes, serving as hard negative examples. The proposals used for both training and evaluation are the top 100 boxes generated by WeDetect-Base-Uni. During training, if a ground truth box has no overlap with any of the proposed boxes, it is added directly to the proposal set to ensure that every ground truth instance has at least one positive match. The input image size is constrained to range from 900 * 32 * 32 to 1600 * 32 * 32, corresponding to 900–1600 visual tokens. Since WeDetect-Ref can process only one query per forward pass, we perform inference separately for each query and then merge the results when evaluating on object detection benchmarks. For example, on the COCO dataset, the model is run 80 times per image, once for each category.

\section{More Experiment Results}

\begin{table}[t]
  \centering
  \caption{Zero-shot evaluation on the D$^3$~\cite{d3} dataset.}
      \begin{tabular}{l|ccc}
        \hline
        Model & Full & Pres & Abs \\
        \hline
        OFA-DOD-B~\cite{d3} & 21.6 & 23.7 & 15.4 \\
        Grounding-DINO-B~\cite{grounding_dino} & 20.7 & 20.1 & 22.5 \\
        OWLv2~\cite{OWL-ST} & 22.8 & 22.1 & 24.7 \\
        GLIP-T~\cite{GLIP} & 19.1 & 18.3 & 21.5 \\
        FIBER-B~\cite{fiber} & 22.7 & 21.5 & 26.0 \\
        Gen-Enhanced-Negs~\cite{zhao2024generating} & 26.0 & 25.2 & 28.1 \\
        Weak-to-Strong~\cite{park2024weak} & 30.8 & 31.0 & 30.4 \\
        \hline
        InternVL2-8B~\cite{chen2024expanding} & 9.8 & 11.0 & 6.2 \\
        Griffon-13B~\cite{zhan2024griffon} & 12.3 & 12.4 & 12.2 \\
        Groma-7B~\cite{ma2024groma} & 16.0 & 15.9 & 16.3 \\
        InternVL2-76B~\cite{chen2024expanding} & 25.3 & 25.7 & 23.5 \\
        ROD-MLLM-7B~\cite{yin2025rod} & 29.7 & 30.0 & 28.7 \\
        \hline
        \rowcolor{ours} WeDetect-Ref 2B & 41.8 & 43.9 & 35.4 \\
        \rowcolor{ours} WeDetect-Ref 4B & \textbf{42.0} & \textbf{44.0} & \textbf{35.8} \\
        \hline
      \end{tabular}
  \label{tab:d3}
\end{table}

\subsection{Performance on Language-Based Object Detection}

Language-Based Object Detection expands category names to flexible language expressions for open-vocabulary object detection (OVD) and overcomes the limitation of referring expression comprehension (REC) only grounding the pre-existing object, which is a combination task of them. We evaluate our model on the commonly used D$^3$~\cite{d3} dataset. As shown in \Cref{tab:d3}, WeDetect-Ref significantly outperforms other models by more than 10 AP, thanks to its high performance on both OVD and REC.

\subsection{Detailed WeDetect Evaluation Results}

In \Cref{tab:detailed_coco}, \Cref{tab:detailed_odinw}, and \Cref{tab:detailed_coco_o}, we provide detailed evaluation results of WeDetect on COCO~\cite{coco}, ODinW35~\cite{li2022elevater}, and COCO-O~\cite{mao2023coco}, respectively. The high and balanced performance across all benchmarks demonstrates the superior open-vocabulary capacities of the WeDetect series.

\begin{table}[t]
  \centering
  \caption{Detailed zero-shot results on COCO~\cite{coco}.}
  \resizebox{\linewidth}{!}{
      \begin{tabular}{l|cccccc}
        \hline
        Model & AP & AP$_{50}$ & AP$_{75}$ & AP$_{s}$ & AP$_{m}$ & AP$_{l}$ \\
        \hline
        WeDetect-Tiny & 44.9 & 61.4 & 48.9 & 26.6 & 49.8 & 62.3 \\
        WeDetect-Base & 52.1 & 69.4 & 57.0 & 34.8 & 57.1 & 69.2 \\
        WeDetect-Large & 54.5 & 72.9 & 59.7 & 42.2 & 59.9 & 66.5  \\
        \hline
      \end{tabular}
    }
  \label{tab:detailed_coco}
\end{table}

\begin{table*}[t]
  \centering
  \caption{Detailed zero-shot results on ODinW35~\cite{li2022elevater}.}
  \resizebox{\linewidth}{!}{
      \begin{tabular}{l|cc|cccc}
        \hline
        Dataset & ODinW13 & ODinW35 & WeDetect-Tiny & WeDetect-Base & WeDetect-Large \\
        \hline
        AerialMaritimeDrone large & \checkmark & \checkmark & 10.8 & 13.5 & 11.0 &  \\
        AerialMaritimeDrone tiled  &  & \checkmark & 19.2 & 19.7 & 25.4 &  \\
        AmericanSignLanguageLetters &  & \checkmark & 2.7 & 4.5 & 5.1 &  \\
        Aquarium & \checkmark & \checkmark & 22.6 & 30.0 & 33.3 &  \\
        BCCD &  & \checkmark & 5.2 & 4.0 & 9.4 &  \\
        boggleBoards &  & \checkmark & 0.6 & 0.5 & 1.2 &  \\
        brackishUnderwater &  & \checkmark & 4.2 & 5.8 & 7.5 &  \\
        ChessPieces &  & \checkmark & 6.8 & 3.9 & 11.9 &  \\
        CottontailRabbits & \checkmark & \checkmark & 78.3 & 78.8 & 79.1 &  \\
        dice &  & \checkmark & 2.0 & 1.4 & 2.5 &  \\
        DroneControl &  & \checkmark & 2.9 & 1.7 & 5.3 &  \\
        EgoHands generic & \checkmark & \checkmark & 58.2 & 62.4 & 65.0 &  \\
        EgoHands specific &  & \checkmark & 17.7 & 21.2 & 20.3 &  \\
        HardHatWorkers &  & \checkmark & 7.2 & 9.3 & 10.3 &  \\
        MaskWearing &  & \checkmark & 1.2 & 1.8 & 2.4 &  \\
        MountainDewCommercial &  & \checkmark & 33.0 & 41.9 & 44.2 &  \\
        NorthAmericaMushrooms & \checkmark & \checkmark & 39.8 & 55.0 & 41.3 &  \\
        openPoetryVision &  & \checkmark & 0.0 & 0.2 & 1.1 &  \\
        OxfordPets by breed &  & \checkmark & 0.7 & 1.9 & 2.6 &  \\
        OxfordPets by species & & \checkmark & 1.9 & 11.8 & 8.9 &  \\
        PKLot &  & \checkmark & 0.0 & 0.0 & 0.8 &  \\
        Packages & \checkmark & \checkmark & 67.9 & 68.3 & 71.3 &  \\
        PascalVOC & \checkmark & \checkmark & 61.1 & 64.9 & 66.6 &  \\
        pistols & \checkmark & \checkmark & 52.1 & 65.2 & 69.5 &  \\
        plantdoc &  & \checkmark & 1.7 & 2.2 & 4.2 &  \\
        pothole & \checkmark & \checkmark & 13.6 & 24.5 & 24.3 &  \\
        Raccoons & \checkmark & \checkmark & 53.1 & 50.9 & 52.3 &  \\
        selfdrivingCar &  & \checkmark & 8.2 & 9.3 & 10.4 &  \\
        ShellfishOpenImages & \checkmark & \checkmark & 45.7 & 59.9 & 60.7 &  \\
        ThermalCheetah &  & \checkmark & 0.8 & 0.1 & 2.9 &  \\
        thermalDogsAndPeople & \checkmark & \checkmark & 40.7 & 53.5 & 52.8 &  \\
        UnoCards &  & \checkmark & 0.0 & 0.0 & 1.1 &  \\
        VehiclesOpenImages & \checkmark & \checkmark & 59.9 & 63.3 & 67.3 &  \\
        WildfireSmoke &  & \checkmark & 14.0 & 25.4 & 24.3 &  \\
        websiteScreenshots &  & \checkmark & 3.3 & 3.6 & 5.1 &  \\
        \hline
        ODinW13 Average &  &  & 46.4 & 53.1 & 53.4 &  \\
        ODinW35 Average &  &  & 21.1 & 24.6 & 25.8 &  \\
        \hline
      \end{tabular}
    }
  \label{tab:detailed_odinw}
\end{table*}

\begin{table*}[t]
  \centering
  \caption{Detailed zero-shot results on COCO-O~\cite{mao2023coco}.}
      \begin{tabular}{l|cccccc|c}
        \hline
        Model & Cartoon & Handmake & Painting & Sketch & Tattoo & Weather & Average \\
        \hline
        WeDetect-Tiny & 43.5&  28.8&  42.9&  44.7&  27.1&  44.7& 38.6 \\
        WeDetect-Base & 52.7 & 37.4 & 52.0 & 50.7 & 22.4 & 49.6 & 44.1 \\
        WeDetect-Large & 53.8 & 37.4 & 51.5 & 50.5 & 35.7 & 52.5 & 47.0 \\
        \hline
      \end{tabular}
  \label{tab:detailed_coco_o}
\end{table*}

\section{Visualization}

\begin{figure*}[t]
  \centering
  \includegraphics[width=1\linewidth]{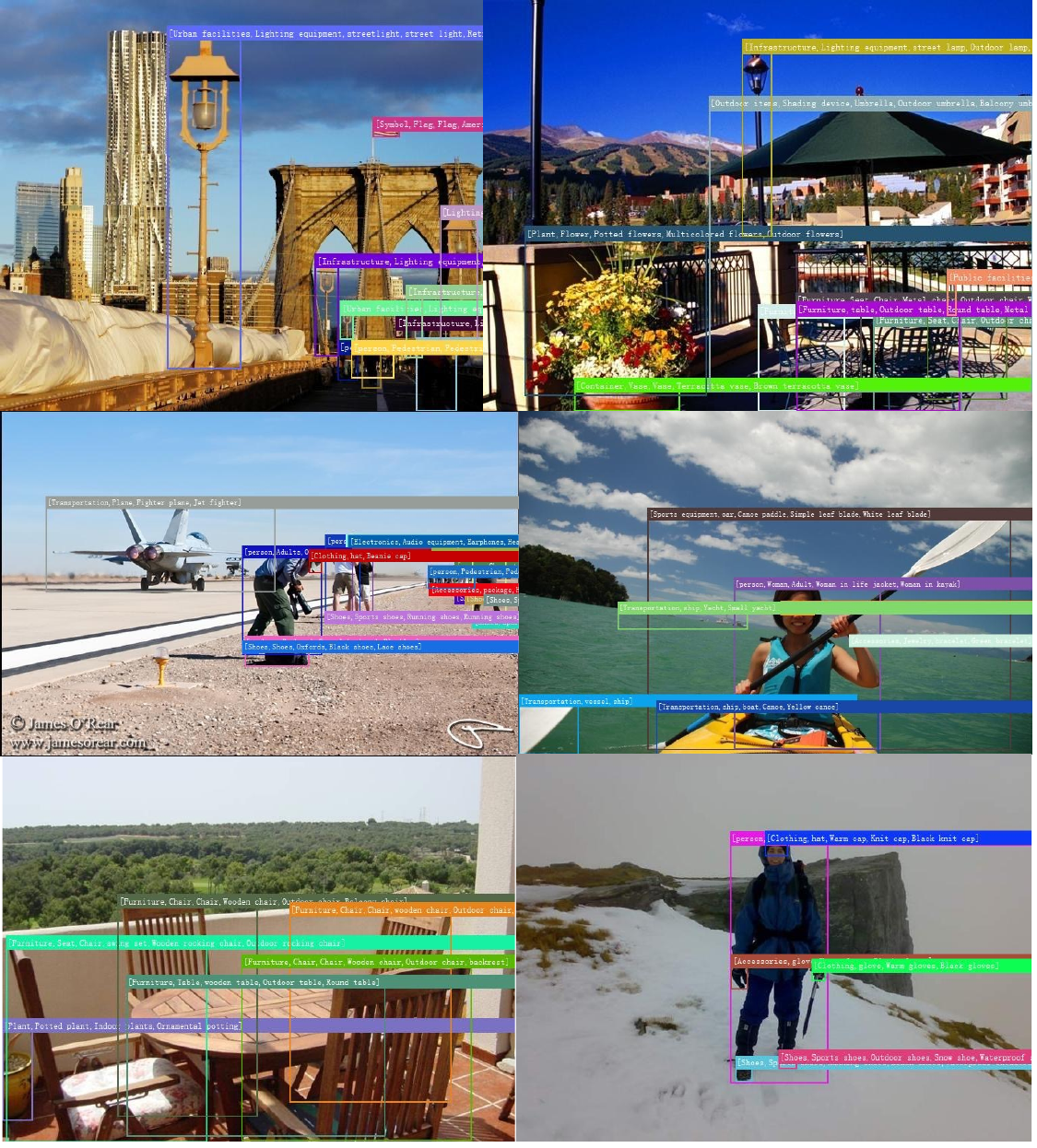}
  \caption{Visualizations of self-annotated data. Each object is annotated with hierarchical labels.}
  \label{fig:anno}
\end{figure*}

\begin{figure*}[t]
  \centering
  \includegraphics[width=1\linewidth]{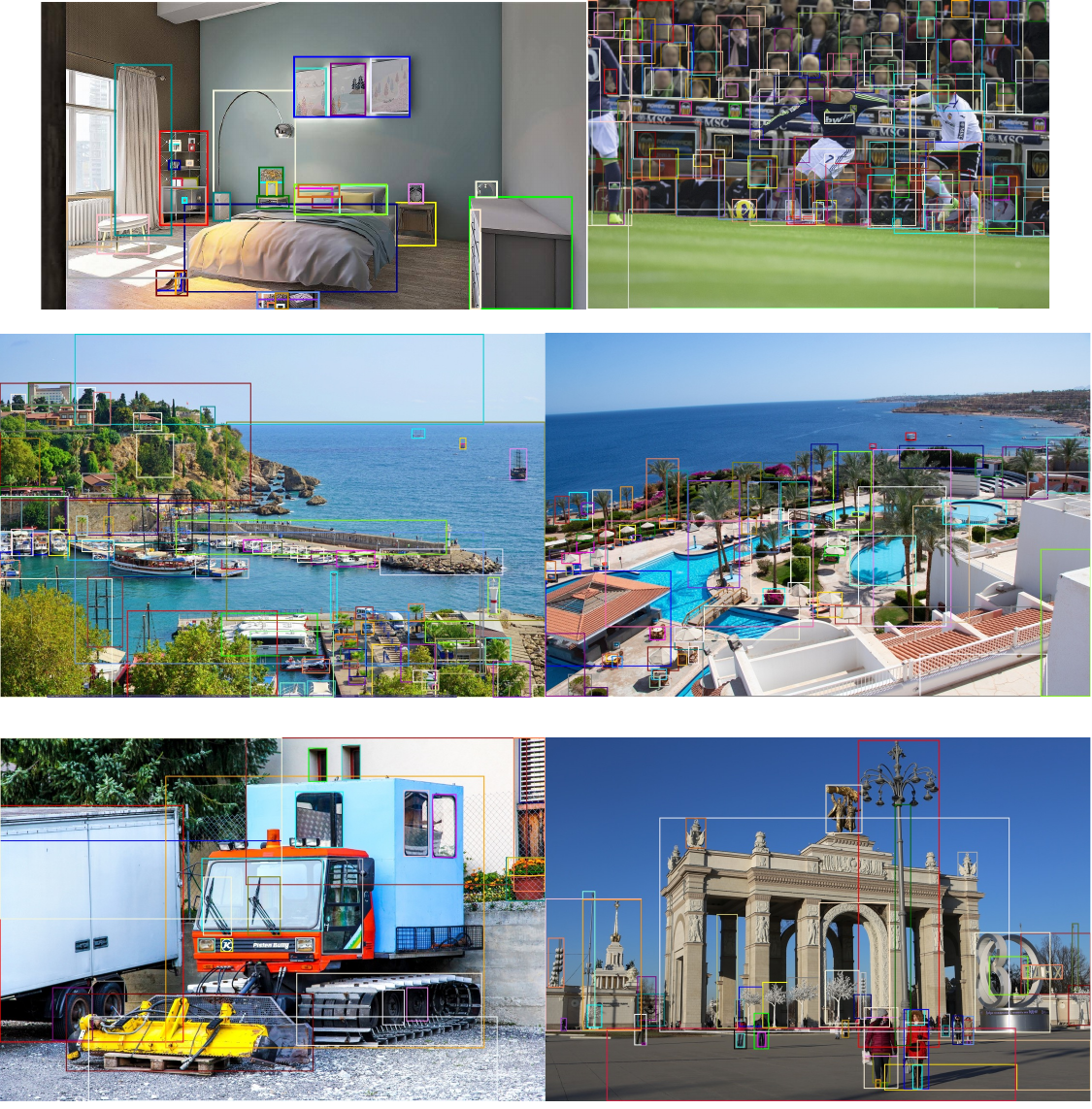}
  \caption{Visualizations of top-scoring proposals generated by WeDetect-Large-Uni.}
  \label{fig:proposals1}
\end{figure*}

\begin{figure*}[t]
  \centering
  \includegraphics[width=1\linewidth]{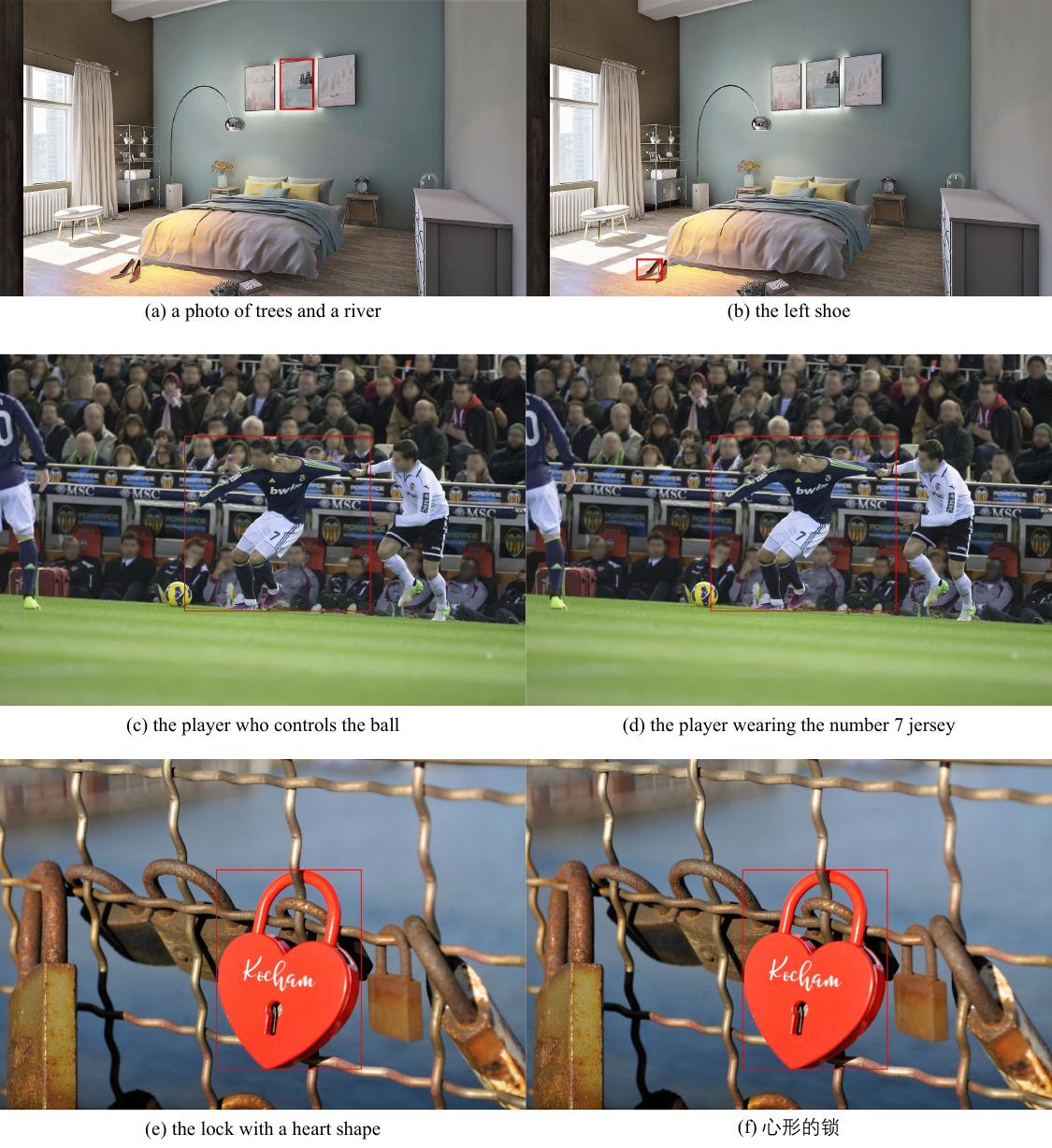}
  \caption{Visualizations of referring expression comprehension results of WeDetect-Ref 4B.}
  \label{fig:ref_example}
\end{figure*}

\subsection{Visualizations of Self-Collected Dataset}

In this work, we develop a data engine to curate a high-quality dataset characterized by balanced concepts, exhaustive annotations, and multi-granularity labels. Examples are shown in \Cref{fig:anno}. For example, in the first picture, the streetlight will be annotated with a hierarchical label list ``[Urban facilities, Lighting equipment, streetlight]''.

\subsection{Visualizations of Inference Results of WeDetect-Uni} 

In \Cref{fig:proposals1}, we visualize some top-scoring proposals extracted by WeDetect-Large-Uni. The model can extract proposals containing both the whole objects and the main parts of the objects with a high recall rate.

\subsection{Visualizations of Inference Results of WeDetect-Ref} 

In \Cref{fig:ref_example}, we visualize several referring expression comprehension examples. We demonstrate that WeDetect-Ref is capable of handling: (a) queries with compositional descriptions, (b) queries involving spatial directions, (c) queries requiring high-level semantic understanding, and (d) OCR-related queries. Interestingly, although WeDetect-Ref is trained exclusively on English data, it can still process multilingual queries (f), thanks to the strong multilingual foundation provided by Qwen3-VL~\cite{Qwen3-VL}.

\end{document}